\PassOptionsToPackage{dvipsnames}{xcolor}
\documentclass[runningheads]{llncs}

\usepackage{graphicx}

\usepackage{tikz}
\usepackage{comment}
\usepackage{amsmath,amssymb} 
\usepackage{color}
\usepackage{array}
\usepackage{ makecell}
\usepackage[pagebackref,breaklinks,colorlinks]{hyperref}

\makeatletter
\newcommand{\printfnsymbol}[1]{%
  \textsuperscript{\@fnsymbol{#1}}%
}
\makeatother

\usepackage[T1]{fontenc}
\usepackage[utf8]{inputenc}
\usepackage{babel}
\usepackage[font=small,labelfont=bf]{caption}
\usepackage{booktabs,hhline}
\usepackage{algpseudocode}
\usepackage[ruled,vlined]{algorithm2e}

\SetCommentSty{mycommfont}

\SetAlgoCaptionSeparator{.}

\usepackage[accsupp]{axessibility}  
\newcommand{\etal}{\textit{et al.}}
\newcolumntype{H}{>{\setbox0=\hbox\bgroup}c<{\egroup}@{}}

\begin{document}
\pagestyle{headings}
\mainmatter
\title{Barlow constrained optimization \\
for Visual Question Answering} 
\titlerunning{Barlow constrained optimization for Visual Question Answering}

\author{Abhishek Jha\thanks{equal contribution}\inst{1} \and Badri N. Patro\printfnsymbol{1}\inst{1} \and Luc Van Gool\inst{1,2} \and Tinne Tuytelaars\inst{1}}
\authorrunning{Jha and Patro et al.}
%

\institute{KU Leuven\inst{1}, 
ETH Z\"{u}rich\inst{2}}
\maketitle
\sloppy
\vspace{-2.2em}
\begin{abstract}
   Visual question answering is a vision-and-language multimodal task, that aims at predicting answers given samples from the question and image modalities. Most recent methods focus on learning a good joint embedding space of images and questions, either by improving the interaction between these two modalities, or by making it a more discriminant space. However, how informative this joint space is, has not been well explored. 
   In this paper, we propose a novel regularization for VQA models, Constrained Optimization using Barlow's theory (COB), that improves the information content of the joint space by minimizing
   the redundancy.
   It reduces the correlation between the learned feature components and thereby disentangles semantic concepts. Our model also aligns the joint space with the answer embedding space, where we consider the answer and image+question as two different `views' of what in essence is the same semantic information.
   We propose a constrained optimization policy to balance the categorical and redundancy minimization forces. When built on the state-of-the-art GGE model, the resulting model
   improves VQA accuracy by $1.4\%$ and $4\%$ on the VQA-CP v2 and VQA v2 datasets respectively. The model also exhibits better interpretability.
\end{abstract}
\vspace{-3.2em}
\section{Introduction}\label{sec:intro}
\vspace{-0.8em}
Visual question answering (VQA) \cite{antol2015vqa} is a challenging vision-and-language task. It involves reasoning about a visual scene based on a free-form natural language question. Answering the question requires learning semantic associations between concepts across the two modalities. As different questions and images referring to the same kind of query and scene should yield a similar answer, learning semantics in the individual modalities and their cross-modal interactions is essential for solving VQA. Many recent works approach this by learning a joint embedding space \cite{Gao_NIPS2015,Ma_AAAI2016,Ren_NIPS2015} or by modeling an attention mechanism \cite{fukui2016multimodal,kim_NIPS2018bilinear,Yang_CVPR2016,Yu_2019_CVPR} in one modality conditioned upon  the other. Another line of work tries to improve the discriminant power \cite{kant2021contrast} of the joint embedding space to improve the answering performance. These have been important contributions. 

\begin{figure}
    \centering     
    \includegraphics[width=0.68\linewidth]{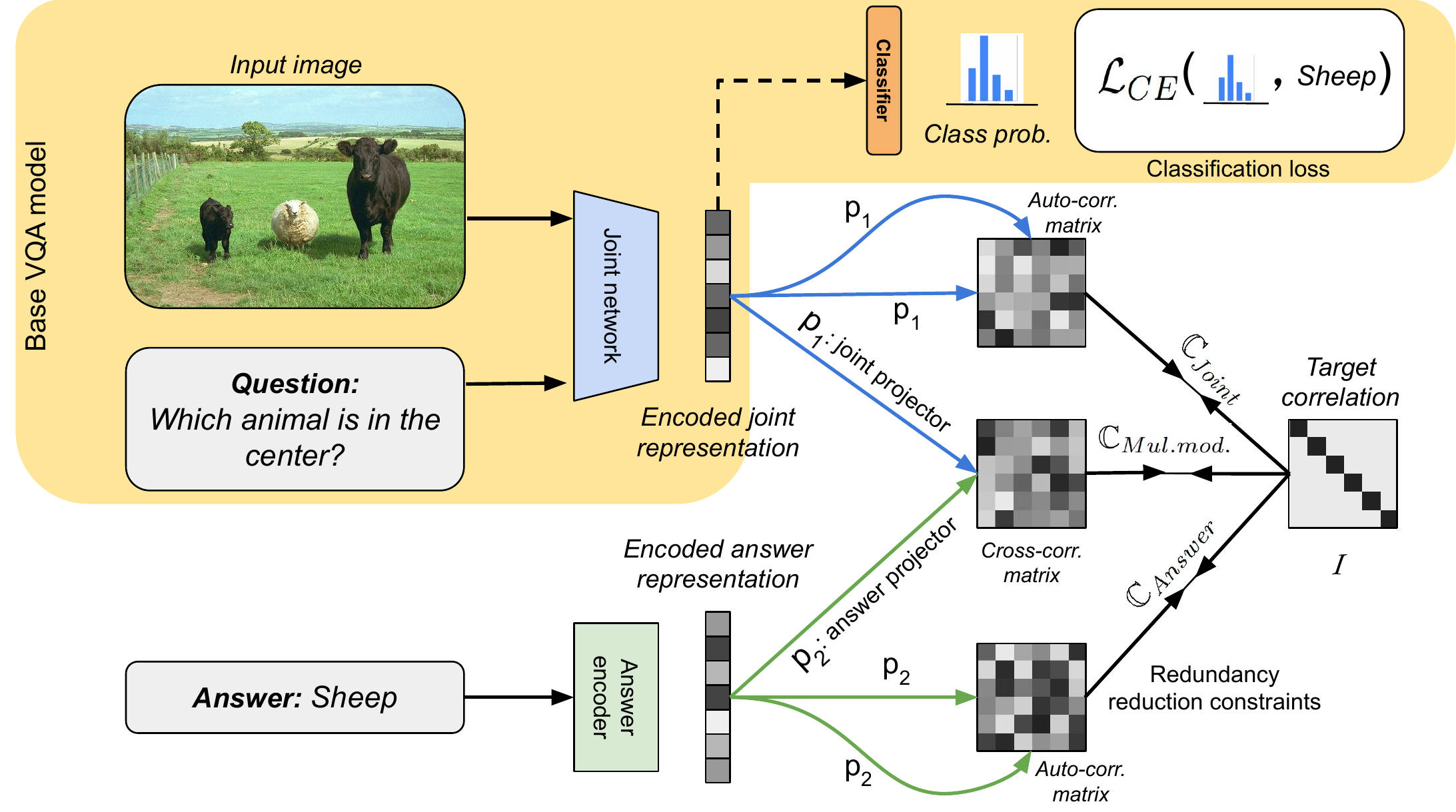}
    \vspace{-0.92em}
    \caption{\textbf{COB}: Given a question and an image, we propose a set of redundancy minimization constraints $\mathbb{C}$ which are applied along with a cross-entropy loss $\mathcal{L}_{CE}$. Unlike existing works, we also incorporate an answer embedding into our VQA system to impose a multimodal redundancy minimization loss. The three constraints minimize the distance between the correlation matrices and the Identity matrix ($I$).}
    \label{fig:teaser}
    \vspace{-2.3em}
\end{figure}

However, high discriminant power of a feature space does not imply high information content \cite{kalantidis2021tldr}. While a highly discriminant space may yield better performance on a loss-specific task by modelling the most discriminant features for a given data distribution, it may be more susceptible to changes in that data distribution. A discriminant space subjected to an additional information preserving constraint, on the other hand, may yield a richer feature space that can generalize better to previously unseen data. 

In this paper, we propose a redundancy reduction constraint, inspired by Barlow's third hypothesis \cite{barlow1961possible} on
sensory message transformations, to incorporate more information in the joint feature space. This third hypothesis (\textit{Redundancy-reducing hypothesis}), states \textit{"that sensory relays recode sensory messages so that their redundancy is reduced but comparatively little information
is lost."}

Redundancy in a feature space arises when multiple feature components cover the same semantic concept. Taking into account the fixed dimensionality of the feature space, this causes the overall information content of the feature space to be suboptimal. A less redundant feature space can model the same information with fewer feature dimensions, or more information with the same number of feature dimensions. This results in a more informative embedding space that model multimodal concepts better and thereby provide a superior VQA performance.

To address this challenge for VQA, we propose an additional decorrelation loss term
besides the categorical loss for predicting the answers. This additional loss term encourages decorrelation across feature components and thereby improves the information content of the embedding space.

Recently for a different task, namely self-supervised representation learning, Zbontar and Jing \etal \cite{zbontar2021barlow}, with their Barlow twins model, have shown that a decorrelation loss, modelled according to  Barlow's Redundancy reduction hypothesis, when applied to two views of the same data offered by the twins model, 
can act as a good supervisory signal to learn visual features. 
Here, we use a similar decorrelation formulation as Barlow twins \cite{zbontar2021barlow}, but
reformulated for two multimodal views of the data.
We pose that the information to be extracted from the image+question input
ideally corresponds to the information present in the answer. In other words, image+question
and answer can be considered as two different 'views' of the same content.
When computing the correlation, we therefore not only consider the auto-correlation
in the joint image+question space, but also the cross-correlation between answer and joint space, as well as the auto-correlation in answer space. As an additional advantage, this brings in information about the semantic similarity between answers via the word embedding used for the answer space. 

Our full pipeline, combining categorical loss with redundancy reduction is shown in Figure \ref{fig:teaser}.

We also found that directly applying the  decorrelation minimization loss (Barlow loss) to a randomly initialized embedding space yields a very high loss. As a result, naively adding a Barlow loss, next to the cross-entropy loss, results in 
inferior VQA results.
On the other hand,  when applying the Barlow loss to the already aligned (pretrained by cross-entropy) embedding space,  this issue does not occur
(see Section \ref{sec:atb_ablation}). Based on this empirical evidence, we formalize a parametric constrained optimization policy to balance the two forces. 
This results in a more informative and discriminant embedding space, leading to an improvement in the answering accuracy.

In summary, our contributions are as follows:

(i) We propose the COB regularization which focuses on redundancy reduction between the joint embedding space of questions and images and the answer embedding space, to improve the information content of a VQA model.

(ii) We propose a policy to balance the categorical and redundancy reduction forces to train the model.

(iii) We improve the  state-of-the-art (SOTA) VQA performance on the challenging VQA v2 \cite{goyal2017making} and VQA-CP v2 \cite{agrawal2018don} datasets.

(iv) Our proposed method
improves the interpretability of the VQA model it builds on.
\vspace{-0.5em}


\vspace{-0.7em}
\section{Related work}
\vspace{-0.9em}
\textbf{Visual question answering : }
VQA has taken up momentum after the introduction of a standard dataset VQA \cite{antol2015vqa} and early multimodal techniques to solve this problem \cite{antol2015vqa,hudson2019gqa,Malinowski_NIPS2014}. Initial approaches \cite{Gao_NIPS2015,Ma_AAAI2016,Ren_NIPS2015} jointly analyze visual and question feature embeddings by concatenating or correlating both features.  In later works \cite{fukui2016multimodal,kim_NIPS2018bilinear,Yang_CVPR2016,Yu_2019_CVPR}, it was observed that attending to specific parts in the images and questions helps to better reason and answer.
The subsequent discovery of language bias in the standard VQA dataset led towards less biased datasets and more robust models.
Agrawal \etal \cite{agrawal2018don} proposed VQA-CP v1 and VQA-CP v2 to overcome the language and distributional bias of the VQA v1 \cite{antol2015vqa} and v2 \cite{goyal2017making} datasets. A critical reasoning-based method proposed by Wu \etal~\cite{wu2019self} ensured the correct answers match the most influential visual regions to overcome the dataset bias. Various authors such as  Ramakrishnan \etal ~\cite{ramakrishnan2018overcoming} proposed an adversarial-based method,
and Jing \etal ~\cite{jing2020overcoming} decomposed a linguistic representation technique to overcome language prior in VQA. Clark \etal~\cite{clark2019don} proposed an ensemble based method to avoid known dataset bias, while Han \etal~\cite{han2021greedy} proposed a gradient ensemble method to overcome both shortcut bias and distributional bias in the dataset. Hence, most methods focus on regularisation techniques to overcome language bias. In this paper, we focus on a regularisation technique to reduce redundancy in the VQA model, and show this further improves its performance.

\textbf{Redundancy reduction:}
Dimensionality reduction is one way of reducing redundancy of a feature space, i.e. by minimizing the number of feature components required to model the data. Linear dimensionality reduction techniques like Principal Component Analysis (PCA) \cite{pearson1901liii} for a single modality provide a mapping between the original feature space and the space spanned by principal components. In this new space low energy principal components can be dropped with a minimal loss of information. Similarly for multiple modalities, Canonical Correlation Analysis (CCA)-like techniques \cite{hardoon2004canonical,hotelling1992relations} provide a linear mapping between individual modalities and a smaller joint embedding space. After CCA, the projections of the modalities are
highly correlated, but they are decorrelated across the resulting feature dimensions.
On the other hand, a manifold with local neighborhood consistency requires non-linear mapping functions \cite{roweis2000nonlinear,zhang2004principal} to preserve the local structure under dimensionality reduction. Our proposed method promotes the learning of decorrelated feature components similar to PCA and CCA. However, unlike PCA and CCA, the learned projection between the original features and the decorrelated component space is non-linear.

Recently, Kalantidis \etal \cite{kalantidis2021tldr} proposed a twin-loss similar to the Barlow twin loss of Zbontar and Jing \etal \cite{zbontar2021barlow} to learn a non-linear dimensionality reduction, as an alternative to PCA. They train a twin encoder-decoder architecture with a decorrelation optimization between the output projections of the nearest neighbors in the input space. Our method is similar to \cite{kalantidis2021tldr} in the way our constraint is motivated, however it is not the primary objective function in our model. We optimize a cross-entropy loss to maximize the answering accuracy, with the decorrelation loss as an optimization constraint. 

\textbf{Decorrelation loss:}
Decorrelation losses are often used in recent representation learning methods \cite{bielak2021graph,kalantidis2021tldr,zbontar2021barlow} by using a shared twin encoder-decoder architecture on two views of the same
samples coming from a unimodal space, while minimizing the distance between an identity matrix and the correlation matrix computed on the output representations.
This forces the feature components in the output embedding space to be orthogonal.
In our case, the inputs come from two different modalities, and hence it differs from the twins formulation. The hypothesis behind the use of our proposed constraint on two different modalities is motivated by the assumption that image-question pairs and their answers should be related to the same underlying concept.

Radford \etal \cite{radford2021learning} propose a similar diagonalization formulation, called CLIP loss, which resembles that of Barlow twins, except it is minimized over different samples within a batch from two different modalities rather than over different dimensions. While the Barlow twins formulation focuses on the diagonalization of the feature correlation matrix, CLIP relies on a cross-entropy loss which forces different samples in the batch to be more discriminative. Our approach also utilizes a multimodal diagonalization formulation, however in the purview of redundancy reduction rather than discriminability.

\textbf{Stabilizing losses: }
Optimizing networks for different objectives requires balancing or weighting the loss gradients, especially for objectives which are non-complementary \cite{achille2019task2vec,higgins2016beta,kingma2013auto,zamir2018taskonomy}, as non-complementary objectives force the feature space to sway in two different directions \cite{higgins2016beta,kingma2013auto}. Improper balancing of such optimization function can lead to trivial solutions \cite{higgins2016beta,rezende2018taming} and hence the loss weighting factor is an important hyperparameter. Rezende and Viola \cite{rezende2018taming} propose a generalized ELBO loss with constrained optimization (GECO), a learnable weighting scheme for balancing KL divergence and the reconstruction loss in the context of training variational auto-encoders. We propose a similar constrained optimization formulation for the cross-entropy loss in our approach, which assigns a dynamic weight to the constraint. Unlike GECO, our objective function and constraint do not have similar scales, with the initial constraint loss being orders of magnitude larger than the optimization function.

\begin{figure*}
    \centering    
    \vspace{-1.5em}

    \includegraphics[width=\linewidth
    ]{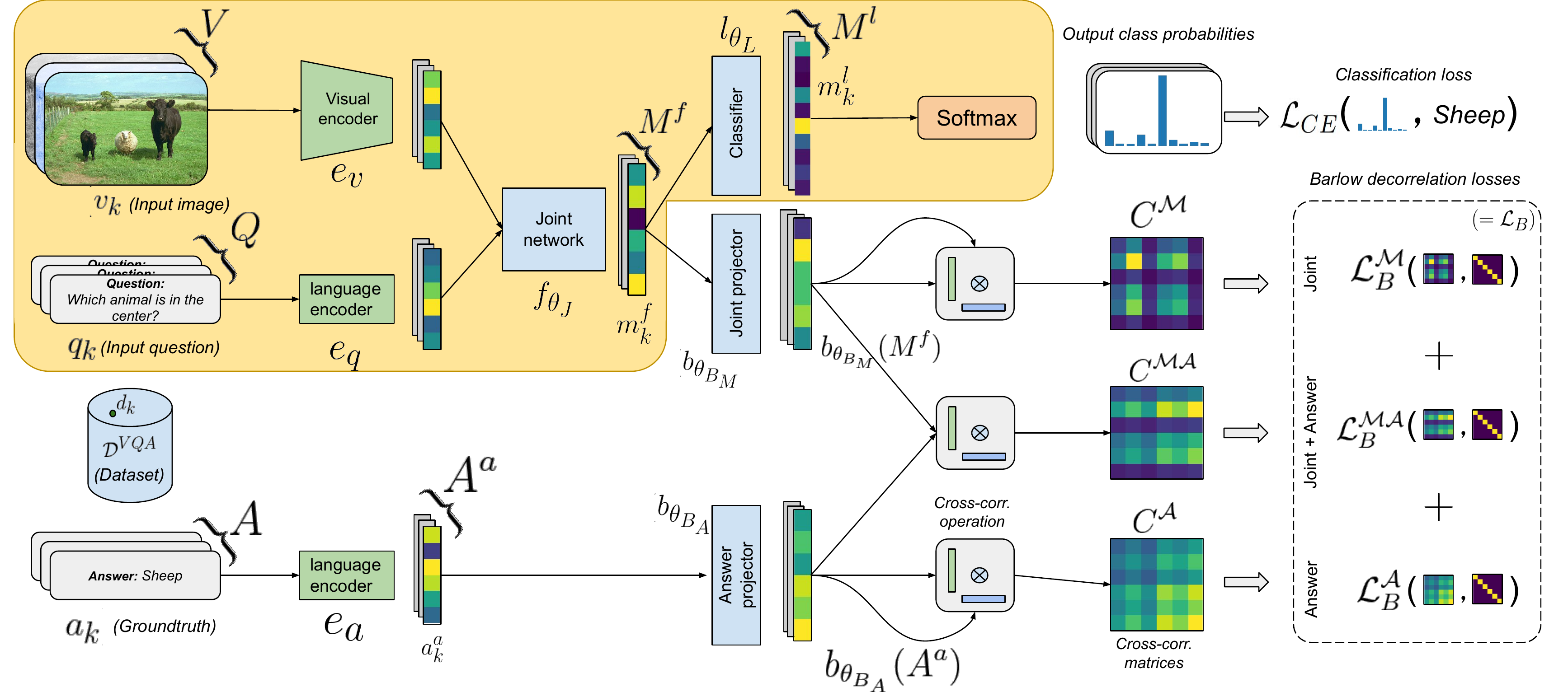}
    \vspace{-1.5em}
    \caption{\textbf{Overall model}: We present the overall COB model, with both classification loss and Barlow redundancy reduction constraint. We explain notations and corresponding components in detail, in section \ref{sec:preliminaries}. We also provide a glossary of all the used notations in the supplementary. All the $\theta$ parameters are learned, while the encoders $\{e_v,e_q,e_a\}$ are pretrained models. During evaluation we only use the classification stream (in yellow) and drop the joint and answer projectors.}
    \label{fig:main_fig}
    \vspace{-4em}
\end{figure*}


\section{Method}\vspace{-0.5em}
\subsection{Preliminaries} 
\label{sec:preliminaries} 
\textbf{VQA formulation:}
The VQA task with cross-entropy loss can be defined as modelling the categorical distribution over a fixed answer vocabulary given a set of image and question pairs. For a data distribution $\mathcal{D}$ for this problem with an instance $d_k = \{v_k, q_k, a_k\} \in \mathcal{D}^{VQA}$, the task is to predict an answer $a_k \in \mathcal{D}^{A}$, given an image $v_k \in \mathcal{D}^{V}$ based on a question $q_k \in \mathcal{D}^{Q}$. Contemporary methods \cite{antol2015vqa,Gao_NIPS2015,Ma_AAAI2016,Ren_NIPS2015} solve this task by first encoding each of the two modalities using pretrained encoders $e_v, e_q$, and then learning a joint representation between them. Each instance pair $(v_k,q_k)$ can then be represented by a point $m^f_k \in \mathcal{D}_M$ in this joint representation space:
\vspace{-0.5em}
 \begin{align}
    m^f_k = {f}_{\theta_J}\big(e_v(v_k),e_q(q_k)\big) \label{eq:m_f_k}\\ 
    m^l_k = l_{\theta_L}(m^f_k) \label{eq:m_l_k}
\end{align}
%
where $f_{\theta_J}$ is the joint network with parameters $\theta_J$, and $l_{\theta_L}$ with parameters $\theta_L$ is the logistic projection, which along with the softmax non-linearity, is used to predict the probability distribution over the answer space $\mathcal{D}^{A}$. A cross-entropy loss ($\mathcal{L}_{CE}$) between the resulting probability scores and the ground truth answer $a_k$ is then computed. For a batch: $(V, Q, A, M^f, M^l)$ consisting of $n_{b}$ number of samples $(v_k,q_k,a_k,m^f_k,m^l_k)$, the cross-entropy loss can be defined as:
\vspace{-1em}
\begin{equation}
  \label{eq:L_CE_1}
    \vspace{-0.5em}
  \begin{gathered}
    \mathcal{L}_{CE}(M^l,A) = -\frac{1}{n_b}\sum^{n_b}_{k}{log\bigg(\frac{\exp(m^l_k[a_k])}{\sum_{a' \in \mathcal{D}^{A}}{\exp(m^l_k[a'])}}\bigg)}
  \end{gathered}
\end{equation}
 where $m^l_k[a_k]$ is the logit corresponding to the answer $a_k$. The resulting gradient is then used to train the parameters of the VQA network.
 
\textbf{Barlow twins formulation:}
In order to reduce redundancy among the feature components, Zbontar and Jing \etal \cite{zbontar2021barlow} propose a distance minimization loss between an identity matrix (\(I\rightarrow\mathbb{R}^{N_B\times N_B}\)) and the correlation matrix (\(C \in \mathcal{D}^{B}\times \mathcal{D}^{B}\)) computed between the non-linear projections \(b_{\theta_B}( . )\) of the encoded representation \(e_{s}(s_k)\) of a positive pair of input samples \(s_{k_1}\) and \(s_{k_2} \in \mathcal{D}^{S}\). For a batch of \(n_b\) such samples \(S = \{s_k\}_k^{n_b}\):
\vspace{-0.7em}
\begin{align}
    &S^{b} = b_{\theta_B}(e_{s}(S)) \label{eq:sbk}\\
    &C(S^{b}_{1},S^{b}_{2}) = Norm_b(S^{b}_{1}) \otimes Norm_b(S^{b}_{2}) \label{eq:C}
\end{align}
where $e_{s}$ is the modality specific feature encoder,  $b_{\theta_B}$ is the non-linear projector from the encoded feature space to a $N_B$ dimensional Barlow optimization space $\mathcal{D}^{B}$, while $Norm_b(.)$ is a normalization function along the batch \cite{ioffe2015batch}. Each element of the correlation matrix $C^\mathcal{S}=C(S^{b}_{1},S^{b}_{2})$ can be indexed by $(i,j)$, as $C^S_{ij}$:
\vspace{-0.5em}
\begin{align}
    &C^\mathcal{S}_{ij} = \frac{\sum^{n_b}_k s^b_{k_1}[i]s^b_{k_2}[j]}{\sqrt{\sum^{n_b}_k (s^b_{k_1}[i])^2}\sqrt{\sum^{n_b}_k (s^b_{k_2}[j])^2}}\\
    &\mathcal{L}^\mathcal{S}_{B} = \sum_{i}^{N_B}{(1-C^\mathcal{S}_{ii})} + \gamma\sum_{i}^{N_B}{\sum_{j}^{N_B}{C^\mathcal{S}_{ij}}} \label{eq:L_B_1}
\end{align}
\noindent where $0<i,j<N_B$ indexes the feature components of the $k^{th}$ sample ($s^b_k \in \mathcal{D}^{B} $) in the projected batch ($S^b:S^b=\{s^b_k\}_k^{n_b}$). The first term in equation \ref{eq:L_B_1}, minimizes the distance between the two projected representations while the second term promotes decorrelation across the feature components, with $\gamma$  a positive hyperparameter to weight the two loss terms.

Our goal is to learn a discriminative space $\mathcal{D}_{M}$, that minimizes $\mathcal{L}_{CE}$ while reducing the redundancy, by reformulating the unimodal $\mathcal{L}^\mathcal{S}_B$ for a multimodal input space ($\mathcal{D}^M, \mathcal{D}^A$). In the next subsections we discuss the different modules of our proposed model in detail.

 \vspace{-0.3em}
\subsection{Objective function formulation}
A typical classification based VQA task can be modelled with equations \ref{eq:m_f_k} to \ref{eq:L_CE_1}. Different methodological improvements have emerged either in the base encoders ($e_v, e_q$), the multimodal interaction between vision and language ($f_{\theta_J}$), or the reasoning network ($l_{\theta_L}$) over the joint embedding. 
We use Greedy Gradient Ensemble (GGE) \cite{han2021greedy} as our baseline and use it as our backbone VQA model.
The GGE-DQ method optimizes both distribution bias and question shortcut bias. It first optimizes a loss between the logit value of a question-only model with the gradient of distributional bias, and then in a second stage, it obtains a loss between the answer logit of the VQA model with a gradient of both distribution bias and question short-cut bias, as discussed in eq.~16 in \cite{han2021greedy}. The joint network of the GGE model can be approximated as $f_{\theta_J}$. Hence our objective function to optimize is the cross-entropy ($\mathcal{L}_{CE}$) in eq.~\ref{eq:L_CE_1}.
\vspace{-0.5em}
\subsection{Baseline model with Barlow loss}\vspace{-0.3em}
First, we combine the cross-entropy objective function $\mathcal{L}_{CE}$ with a decorrelation loss, 
see Fig.~\ref{fig:main_fig}. For a set of encoded question and image representations, $e_q(Q)$ and $e_v(V)$, we obtain a joint representation $M^f$ using eq.~\ref{eq:m_f_k}. This joint representation $M^f$ becomes one of the two modalities which we want to decorrelate. The second modality is the answer space encoded by $A^a = e_a(A)$. We then compute three decorrelation losses: unimodal joint embedding loss $\mathcal{L}_{B}^\mathcal{M}$, unimodal answer embedding loss $\mathcal{L}_{B}^\mathcal{A}$ and a multimodal embedding loss $\mathcal{L}_{B}^\mathcal{MA}$:
 \vspace{-0.7em}
\begin{align}
    &C^\mathcal{M} = C(b_{\theta_{B_M}}(M^f), b_{\theta_{B_M}}(M^f)) \label{eq:C^M}\\
    &C^\mathcal{A} = C(b_{\theta_{B_A}}(A^a), b_{\theta_{B_A}}(A^a)) \label{eq:C^A}\\
    &C^\mathcal{MA} = C(b_{\theta_{B_M}}(M^f), b_{\theta_{B_A}}(A^a)) \label{eq:C^MA}\\
    &\mathcal{L}_{B}^\mathcal{O} = \bigg\{\sum_{i}^{N_B}{(1-C^\mathcal{O}_{ii})} + \gamma\sum_{i}^{N_B}{\sum_{j}^{N_B}{C^\mathcal{O}_{ij}}}\bigg\}_{\mathcal{O} \in ^\mathcal{\{M,A,MA\}}} \label{eq:L_BO}\\
    &\mathcal{L}_B = \mathcal{L}_{B}^\mathcal{M} + \mathcal{L}_{B}^\mathcal{A} + \mathcal{L}_{B}^\mathcal{MA} \label{eq:L_B_2}
\end{align}
\noindent where $C(.)$ is defined in eq.~\ref{eq:C}. Hence, the overall loss $\mathcal{L}_{all_{base}}$ for our baseline model becomes:
\vspace{-1em}
\begin{align}
  &\mathcal{L}_{all_{base}} = \mathcal{L}_{CE} + \mathcal{L}_B \label{eq:L_all_base}
\end{align}
Here the first loss term is to enforce the discriminative property on the joint features $m^f_k$, while the second term reduces correlation between the feature components in both projected answer space and the joint image-and-question space. The gradient of the loss term $\mathcal{L}_B$ in eq.~\ref{eq:L_all_base}, is backpropagated to update $f_{\theta_J}$, which optimizes its parameters to learn the joint representations $m^f_k$ to become less redundant. This results in a joint embedding space that is discriminant and informative.
 \vspace{-0.5em}
\subsection{Balancing the two losses}\label{cob} 
\begin{figure}
    \centering
      \vspace{-2em}
    \begin{minipage}{0.48\textwidth}
        \centering
        \includegraphics[width=1.0\textwidth]{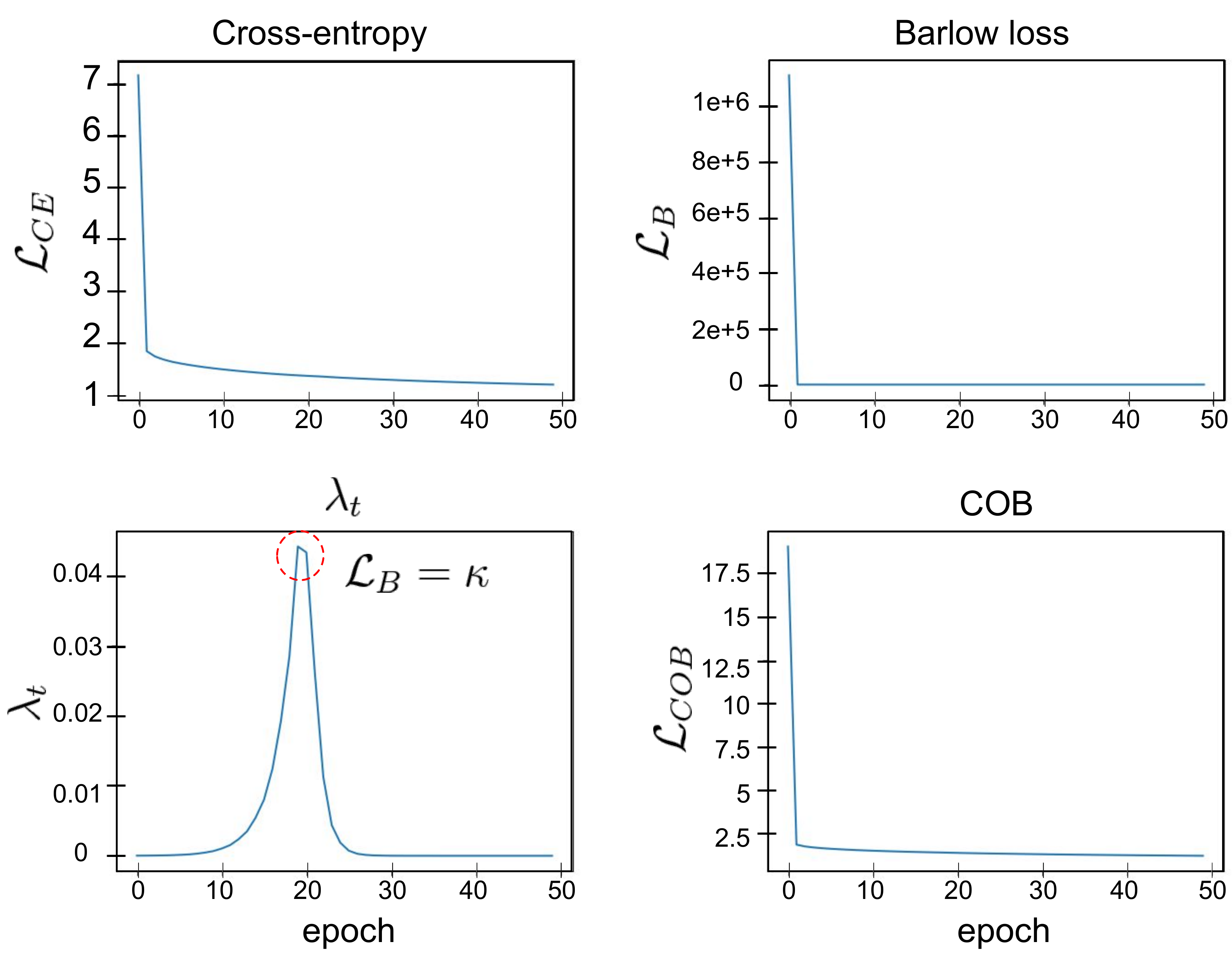} 
        \caption{We plot the loss functions for our COB model during training along with Lagrange multiplier $\lambda$.}
        \label{fig:loss_plot}
    \end{minipage}\hfill
    \begin{minipage}{0.48\textwidth}
        \centering
        \includegraphics[width=1.0\textwidth]{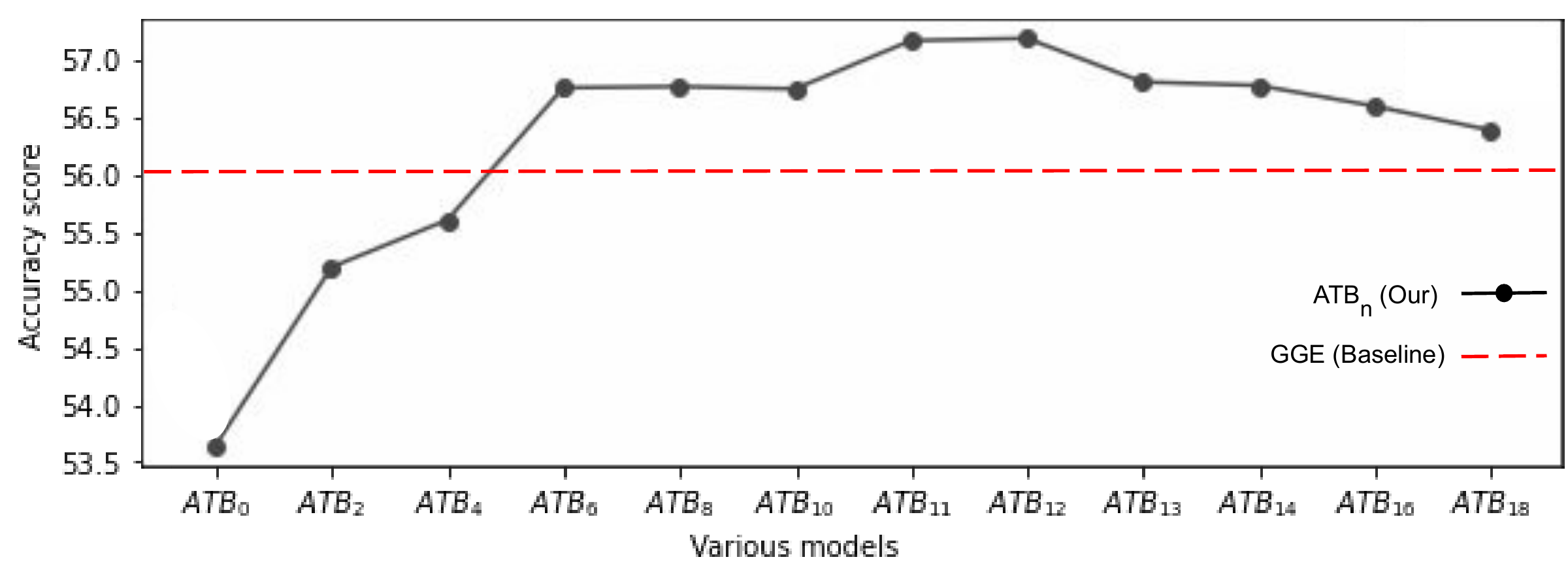} 
        \caption{Ablation analysis: Applying Barlow loss after certain epoch. Individual scores for Y/N, Number and ``Other'' scores are present in supplementary. (In this figure, $A_n$ stands for $ATB_n$, n is the number of pretrained epochs)}
        \label{fig:epoch}
    \end{minipage}
      \vspace{-3.5em}
\end{figure}

\vspace{-0.3em}
Contrary to our initial expectations, 
we observed that, when optimizing the overall loss defined in equation \ref{eq:L_all_base}, the classification performance actually decreases, (see Section \ref{sec:atb_ablation}). We conjecture that the loss in performance occurs because of the
difference in the dynamic range 
of the two loss terms. These losses are non-complementary and promote different properties in the embedding space. While cross-entropy makes the joint embedding space more discriminative, decorrelation tries to make the feature components orthogonal. An optimal weighing of the two loss terms is needed to ensure a rich representation that is discriminative while being informative. We propose two different approaches to achieve this:

\textbf{a) Align then Barlow (ATB):} This is our intermediate model, given to better understand the dynamics between the cross-entropy loss and the decorrelation constraints. In this setup, the VQA network is first pretrained with the cross-entropy loss for $n$ number of epochs and then finetuned with both loss terms, equation \ref{eq:L_all_base}, till the loss converges. The resulting loss $\mathcal{L}_{all_{ATB}}$ can be written as equation \ref{eq:L_all_ATB}. On analysing the Barlow twins \cite{zbontar2021barlow} evaluation loss curve, we observe that the Barlow loss requires a large number of epochs to converge ($\sim1000$). This suggests that the Barlow twins loss surface is flatter requiring more gradient cycles to converge. Therefore a pretraining step to learn a meaningful representation can expedite the convergence as orthogonalization of learned features can be viewed as rotating them in the representation space. In contrast, for a randomly initialized feature space, the network has to learn meaningful features and perform rotation simultaneously.
\begin{align}
    \mathcal{L}_{all_{ATB}}=
    \begin{cases}
      \mathcal{L}_{CE}, & \text{if}\ epoch\leq n \\
      \frac{1}{2}\big(\mathcal{L}_{CE} + \mathcal{L}_B\big), & \text{otherwise}
    \end{cases}\label{eq:L_all_ATB}
\end{align}
\textbf{b) Constrained optimization using Barlow's theory (COB)}: The Barlow decorrelation loss on a randomly initialized joint embedding space is orders of magnitude larger than the cross-entropy, as shown in Figure \ref{fig:loss_plot}. This high imbalance in the losses forces the network to move towards decorrelation optimization, and as discussed before, the decorrelation loss surface is flatter and hence the network does not converge when having a high loss imbalance. However, if the network is pretrained with cross-entropy loss for certain number of epochs, the Barlow decorrelation loss decreases swiftly. This calls for a dynamic weighing scheme which changes based on the degree of imbalance between the two losses. Inspired by \cite{rezende2018taming}, we propose a constrained optimization formulation of equation \ref{eq:L_all_base} to dynamically control the weights assignment to the two loss terms:
\vspace{-1.0em}
\begin{align}
&\mathcal{L}_{all_{COB}} = \mathcal{L}_{CE}; \quad \text{subject to} \quad \mathbb{C}^t \leq 0 \label{eq:L_all_COB}\\
&\mathbb{C}^t = \alpha\mathbb{C}^{t-1} + (1-\alpha)(\mathcal{L}_{B}-\kappa) \label{eq:C_t_form}
\end{align}
where $\mathbb{C}^t$ captures the momentum of Barlow constraint $\mathcal{L}_{B}$ per epoch with $\alpha$ being the momentum factor and $\kappa$ is a tolerance hyperparameter \cite{rezende2018taming}. The above equation \ref{eq:L_all_COB} can be rewritten as a non-constrained optimization problem:
\begin{align}
&\mathcal{L}_{all_{COB^\lambda}} = \mathcal{L}_{CE} + \lambda_{t} \mathbb{C}^t \label{eq:L_all_COB_lambda}\\
& \lambda_t \leftarrow \lambda_{t-1} \exp(\mathbb{C}^t) \label{eq:lambda_t}\\
&\triangle\lambda_t \propto \exp(\mathbb{C}^t) \label{eq:del_lambda_t}
\end{align}
where $\lambda_t$ is the Lagrange multiplier ($\lambda$) at iteration $t$. The change is 
directly proportional to the exponential of the magnitude of the Barlow loss. Here, $\lambda_t$ is initialized with a small value to bring both the loss terms in a similar range. While $\mathcal{L}_{B}$ itself consists of three loss terms, equation \ref{eq:L_B_2}, we use a single value of $\lambda_t$ to weight all of them, as their values vary in a similar range. This simplifies the overall formulation and reduces the number of non-gradient parameters ($\lambda$) to update.

 \vspace{-1.5em}
\section{Experiments}
  \vspace{-0.5em}

\textbf{Evaluation Metric:} 
We use the answering accuracy, the standard evaluation metric for VQA \cite{antol2015vqa}, to evaluate all our models. We use another metric Correctly Grounding Difference (CGD) \cite{han2021greedy}, which is the difference of CGR\cite{shrestha2020negative} (Correct Grounding for Right prediction) and CGW (Correct Grounding but Wrong prediction) to evaluate the visual grounding of a method. To evaluate our proposed model we conduct experiments on the standard VQA v2 \cite{goyal2017making} and language-bias sensitive VQA-CP v2 \cite{agrawal2018don} datasets. We discuss more about the datasets in the supplementary.

\subsection{Training details}
We train our COB model using the classification loss and the Barlow loss in an end-to-end manner. We use GGE-DQ-iter\cite{han2021greedy} as our base model.
To update model parameters, we use the Adamax\cite{kingma2014adam} optimizer with configured hyper-parameter values as follows: \{learning rate = 0.001, batch size = 512, beta = 0.95, alpha = 0.99 and epsilon = 1e-8\}. To train this COB model, we configure the hyperparameters of the constraint formulation as follows:
\{$\lambda_{init}$ = 0.0001, step size = 100, $\kappa= 2.63$ and $\alpha$ = 0.99\}. 
Here $\lambda$ is a learnable parameter, and it updates based on the moving average of the constraint loss as discussed in Section \ref{cob}. We initialize with $\lambda_{init}=0.0001$. The value of $\lambda$ updates after every 100 iterations based on the Barlow constraint loss value. The constraint loss depends on the previous constraint value and current value with a factor of $\alpha=0.99$ and $1-\alpha$ as shown in equation \ref{eq:C_t_form}. Initially, the $\lambda$ value starts increasing, and after the Barlow loss ($\mathcal{L}_B$) reaches the threshold value ($\kappa = 2.63$), it starts decreasing as shown in Figure \ref{fig:loss_plot}. More details about the model architecture are provided in the supplementary.
\vspace{-0.5em}

 \begin{table}[htb]
    \vspace{-2.5em}
  \centering
  \caption{Ablation analysis of our approach}\label{tab:abl}
  \begin{tabular}{@{}lcccccccc@{}}\hline
    Method &$\mathcal{L}_{CE}$&$\mathcal{L}_{B}^\mathcal{M}$&$\mathcal{L}_{B}^\mathcal{MA}$&$\mathcal{L}_{B}^\mathcal{A}$& All & Y/N & Number & Other \\ \hline
    GGE  &\checkmark&&&& 56.08 & 86.64& 22.15&49.38\\
    $COB^\mathcal{M}$ &\checkmark&\checkmark&&& 57.03 & 87.17& 26.67 & 49.57\\
    $COB^\mathcal{MA}$ &\checkmark&&\checkmark&& 56.77 & 86.84 & 24.83 & 49.75\\
    $COB^\mathcal{M,MA}$ &\checkmark&\checkmark&\checkmark&& 57.49 & 86.57 & \textbf{30.12} & \textbf{49.77}\\
    $COB$ &\checkmark&\checkmark&\checkmark&\checkmark& \textbf{57.53} & \textbf{88.36} &28.81 & 49.27\\
    \hline
  \end{tabular}
  \vspace{-3.5em}
\end{table}

\subsection{Ablation: Epoch analysis for ATB}
 \label{sec:atb_ablation}
    
    In this section, we discuss the effect of the pretraining epochs for the ATB model on the final VQA performance. This analysis is critical as it demonstrates that naive addition of the two loss terms, as in equation \ref{eq:L_all_base}, is not the best training policy. Figure \ref{fig:epoch} shows the performance of our ATB model at convergence for different pretraining epochs. Without pretraining, a drop in performance of more than $2\%$ can be observed. When the model is fine-tuned on lesser amount of pretraining ($ n < 11$), the performance is inferior at convergence. As the initial loss of Barlow decorrelation is orders of magnitude higher, and the two loss terms are non-complementary, the resulting gradient for cross-entropy loss is relatively weaker to learn good discriminative features. We also observe that the accuracy increases with increase in pretraining epochs, this happens as the loss for Barlow decorrelation for a pretrained feature space converges faster. Since for a pretrained feature space, decorrelation is analogous to rotating the feature components towards their orthogonal principal axis, the Barlow decorrelation loss finds it easier to converge. This results in gradients for both cross-entropy loss and Barlow decorrelation to be comparable, and hence results in learning a richer feature space. Finally, we see a drop in performance, for a larger pretraining epoch ($n > 12$). For a larger number of pretraining epochs, the validation cross-entropy loss starts to overfit and the non-complementary Barlow decorrelation loss no longer improves the performance.
    

\begin{table*}[htb]
    \centering	
    \caption{SOTA:  VQA-CP v2 accuracy on test-set and VQA v2 accuracy on val set. Methods with * use extra annotations (e.g., human attention (HAT)~\cite{das2016human}, explanations (VQA-X)~\cite{park2018multimodal}, or object label information).
    GGE-iter (impl.) is our implementation of GGE-DQ-iter\cite{han2021greedy} model. We sort Table-\ref{sota} based on VQA-CP v2 scores.}\label{sota}
    \begin{tabular}{l H cccc  c cccc} \hline
    & &\multicolumn{5}{c|}{VQA-CP v2 \cite{agrawal2018don} test} & \multicolumn{4}{c}{VQA v2 \cite{goyal2017making} val} \\ \cline{3-7} \cline{8-11}
		Models                  & Base & All & Y/N & Number & Other     & CGD       & All & Y/N & Number & Other\\ \hline 
		CSS(UpDn)* \cite{chen2020counterfactual}       &UpDn& 41.16& 43.96& 12.78& 47.48& 8.23& 59.21& 72.97& 40.00& 55.13\\
		AdvReg.\cite{ramakrishnan2018overcoming} &UpDn & 41.17 & 65.49 & 15.48 & 35.48  &- &62.75& 79.84& 42.35& 55.16\\
		RUBi \cite{cadene2019rubi}               &UpDn & 45.42 & 63.03 & 11.91 & 44.33  &6.27& 58.19& 63.04& 41.00& 54.43\\
		Hint*\cite{selvaraju2019taking}         &UpDn & 47.50 & 67.21 & 10.67 & 46.80  &10.34 &63.38& 81.18& 42.14& \underline{\textbf{55.66}}\\
		GVQE*\cite{kv2020reducing}              &UpDn& 48.75& -& - &- &- &\textbf{64.04}& -& -& -\\
	 	LM  \cite{clark2019don}                 &UpDn& 48.78& 70.37& 14.24& 46.42& 11.33& 63.26& 81.16& 42.22& 55.22\\
        DLP \cite{jing2020overcoming}           &UpDn& 48.87& 70.99& 18.72& 45.57& -& 57.96& 76.82& 39.33& 48.54\\
        SCR* \cite{wu2019self}                  &UpDn & 49.45 & 72.36 & 10.93 & 48.02  &-& 62.20& 78.8& 41.6& 54.4\\ 
        LMH\cite{clark2019don}                  &UpDn& 52.73& 72.95& \underline{\textbf{31.90}}& 47.79& 10.60& 56.35& 65.06& 37.63& 54.69\\
        CF-VQA\cite{niu2021counterfactual}     &UpDn &53.69& \textbf{91.25}& 12.80& 45.23& -& 63.65& \textbf{82.63}& \textbf{44.01}& 54.38\\ 
        GGE-iter\cite{han2021greedy}         &UpDn &57.12 &87.35 &26.16 &\textbf{49.77}& \underline{\textbf{16.44}} &59.30& 73.63& 40.30 &54.29\\\hline
        GGE-iter (impl.)                     &UpDn  & 56.08 & 86.64& 22.15&\underline{\textbf{49.38}}& 15.92 &58.92 &72.00 & 40.13 & 53.95 \\
        \textbf{COB(ours)}                      &UpDn &\underline{\textbf{57.53}} & \underline{\textbf{88.36}} &28.81 & 49.27 &\textbf{16.89} &\underline{\textbf{63.80}} &\underline{\textbf{81.36}} & \underline{\textbf{43.30}} &\textbf{55.86}\\\hline
        CSS(LMH)*\cite{chen2020counterfactual}  &LMH &\textbf{58.21} &83.65 & \textbf{40.73} &48.14 &8.81 &53.15 &61.20 &37.65 &53.36\\
    \hline
 	\end{tabular}
		\vspace{-1.4em}
\end{table*}

\begin{figure*}
    \centering
    \includegraphics[width=\linewidth]{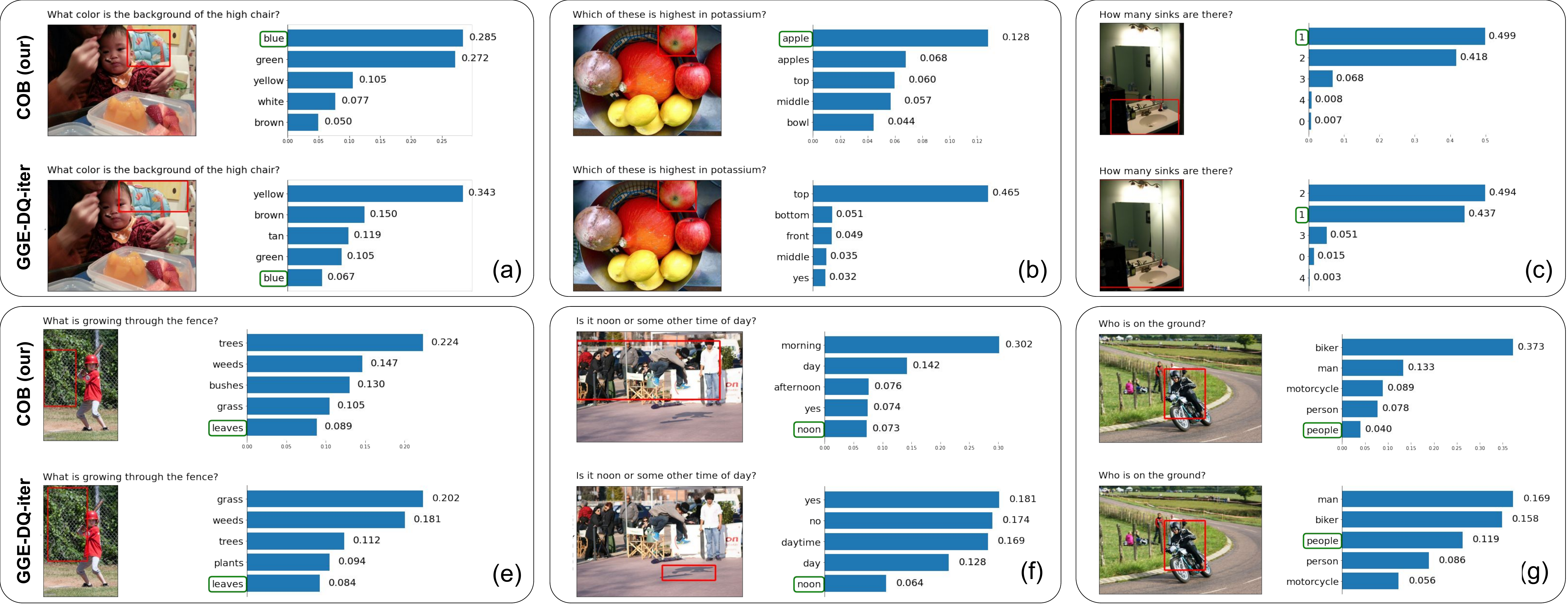}
      \vspace{-1.7em}
      \caption{\textbf{Qualitative results:} Each set of images show the input image-question pair and top-5 predictions for our proposed COB model compared against the baseline GGE-DQ-iter model. {\color{red}Red bounding box} shows the maximal attention region in each image. Answers within the {\color{ForestGreen}green boxes} are the ground truths. We see that COB performs better with higher prediction score to the ground truth answer in comparison to the baseline method (a)-(e). For negative results as well, the predicted classes are semantically relevant. This is further analyzed in the context of the explainability in section \ref{sec:explainability}. We provide more qualitative results in supplementary.}
      \label{fig:result_main}
      \vspace{-2em}
  \end{figure*}

\subsection{Ablation of proposed approaches}\vspace{-0.5em}
Our constraint formulation $\mathcal{L}_B$ consists of three loss terms $\mathcal{L}_{B}^\mathcal{M}$, $\mathcal{L}_{B}^\mathcal{A}$ and $\mathcal{L}_{B}^\mathcal{MA}$, equation \ref{eq:L_B_2}. To understand the importance of each of these loss terms, we ablate them individually in the constraint and re-train the COB model. For the model with only $\mathcal{L}_{B}^\mathcal{M}$ loss, i.e. $COB^\mathcal{M}$, the answering accuracy is $57.03\%$, better than the baseline GGE model, as shown in Table \ref{tab:abl}. This shows that increasing information content (or minimizing the redundancy) of the joint features helps VQA performance. 
$COB^\mathcal{MA}$, that  contains the constraint term $\mathcal{L}_{B}^\mathcal{MA}$, forces the model to learn an alignment between the answer and the joint features in the projected Barlow space while maintaining the decorrelation between the feature components. The gradients from $\mathcal{L}_{B}^\mathcal{MA}$ provide an additional supervision along with $\mathcal{L}_{CE}$ to help the underlying joint embedding space $m^f_k$ to learn features relevant to the answer, resulting in an answering performance of $56.77\%$, Table \ref{tab:abl}. Combining the these two constraint terms, $\mathcal{L}_{B}^\mathcal{M}$ and $\mathcal{L}_{B}^\mathcal{MA}$, in $COB^\mathcal{M, MA}$ results in an increased performance of $57.49\%$. Finally, $COB$ model contains all three loss terms, the additional $\mathcal{L}_{B}^\mathcal{A}$ improves the information content of the answer embedding. This further assists the $\mathcal{L}_{B}^\mathcal{MA}$ loss to learn a better alignment between the less redundant joint and the answer embedding spaces, outperforming the other three ablated baselines. This ablation analysis shows that each of the three loss terms in our constraint provides a different supervision to the model and thereby improves the underlying joint representations.

\vspace{-0.7em}
\subsection{Comparison with State-of-the-art} \vspace{-0.5em}
We provide performance results on two datasets, challenging VQA-CP v2\cite{agrawal2018don} that has a less language bias and a standard VQA v2\cite{goyal2017making} dataset in Table \ref{sota}.  CSS\cite{chen2020counterfactual} \& CF-VQA\cite{niu2021counterfactual} use counterfactual examples to overcome bias,  AdvReg\cite{ramakrishnan2018overcoming} uses regularisation techniques, HINT\cite{selvaraju2019taking} \& SCR\cite{wu2019self} use grounding techniques, RUBi\cite{cadene2019rubi}, LM\cite{clark2019don} and GGE\cite{han2021greedy} use ensemble methods, GVQE \cite{kv2020reducing} \& DLP\cite{jing2020overcoming} use new encoder based method to overcome language and dataset bias.  Some methods use extra annotations to improve debiasing performance, but our method does not use any extra annotations and performs better than most current state-of-the-art (SOTA) methods with better explainability in the results (see Section~\ref{sec:explainability}). Our implementation of GGE model performance is 56.08\% and 58.92\% on VQA-CP v2 and VQA v2 datasets respectively. In comparison our COB model, built upon the base GGE model, obtains a performance of 57.53\% and 63.80\%, which is an improvement of 1.45\% and 4.9\% respectively. We also outperform the official GGE \cite{han2021greedy} performance.
Our COB model outperforms other SOTA methods, that do not use extra annotations, for overall answer prediction task on both VQA-CP v2 and VQA v2 datasets, as presented in Table \ref{sota}. We also improve overall CGD score by 0.45 units, which shows that our model is able to learn a better grounding between vision-and-language modalities.  

\vspace{-1em}
\subsection{Qualitative results}
\label{sec:qual_results}
 Figure \ref{fig:result_main} illustrates top-5 answers and probability scores for a few examples. We compare our qualitative results with the most recent state-of-the-art method GGE-DQ \cite{han2021greedy}. In the first and third examples (part a and c of Figure \ref{fig:result_main}), our COB model attends a more precise salient region leading to a correct answer as compared to GGE-DQ model whose attention region extends over a larger non-salient region, thus answering incorrectly. For the second image, both the models focus on the same region, however COB assigns a higher probability score to the correct answer. These results indicate that the more informative latent features provide better reasoning, improving the localization and the probability scores for the correct answers compared to the baseline method.  Similarly, we show results for various combinations of attention and answer prediction results.

 \begin{figure}
    \centering
    \includegraphics[width=\linewidth]{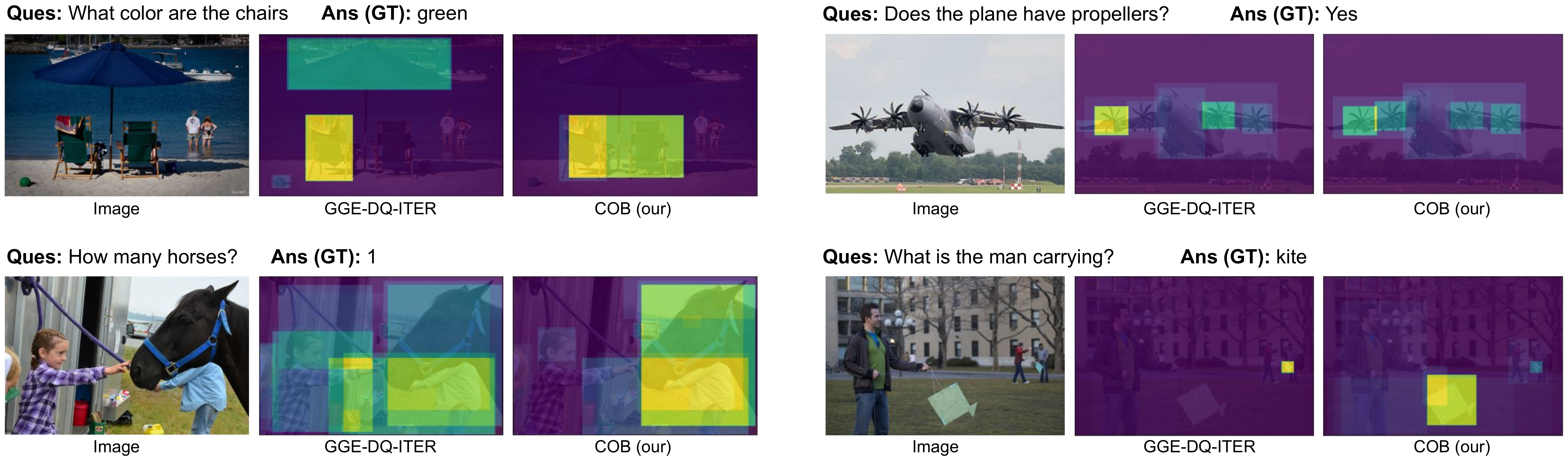}
    \vspace{-2em}
    \caption{\textbf{Explainability of the models:} Given an image and the question, we show the class activation maps for the samples in the joint embedding space $m^j_k$ corresponding to the answer. We observe that the COB model's Grad-CAM outputs are better localized in the salient regions for answering the question. More examples are shown in supplementary.}
      \label{fig:result_cam}
      \vspace{-2.2em}
  \end{figure}

\vspace{-.7em}
\section{Analysis and Discussion}

\vspace{-.5em}
\subsection{Explainability: Grad-Class activation maps}
\label{sec:explainability}
\vspace{-.2em}

Reasoning is an essential part of question answering, and is directly influenced by the quality of the joint representation space.
Hence it is crucial to study what the model has learned and how it processes the input data. This interpretability of a model is even more important for failure cases, in order to understand the cause of failure and model shortcomings.
We use Grad-CAM\cite{Selvaraju_2017_ICCV} as an indicative of model interpretability by computing the saliency for an image given the question and the groundtruth answer.
We analyze our COB and the SOTA baseline model \cite{han2021greedy}, trained on VQA-CP v2, in the context of model interpretability in Figure \ref{fig:result_cam}. We observe that our model produces more interpretable regions compared to the baseline GGE model, which also indicates the reason for a higher CGD score in Table \ref{sota}. For both examples, our model focuses on correct regions that are salient for answer prediction.

  \begin{minipage}{\textwidth}
   \hspace*{-2em}
  \begin{minipage}[t!]{0.45\textwidth}

      \begin{tabular}{@{}l ccccc}
    \hline
     &\multicolumn{4}{c}{\thead{Cumulative energy for \\  top-$k$ PCA components (in\%)}} \\ \cline{2-5}
    $N_B$ & $k=512$ & $k=256$ & $k=128$ & $k=64$ \\
    \hline
    512  & 100 & 99.8& 99.3&94.1\\
    1024 & 99.9& 99.5&98.1 &80.4\\
    2048 & 99.7 & 98.5 & 91.1 & 65.0\\
    4096 &98.8 &95.7 &85.2 & 59.9\\
    \hline
  \end{tabular}
      \captionof{table}{\textbf{Projector dimensionality ($N_B$) selection:} PCA energy for top-\(k\) components for different Barlow projection ($b_{\theta_B}$)  dimensions.}
  \label{tab:PCA_proj}
      \end{minipage}
    \hspace*{0.5em}
  \begin{minipage}[t!]{0.5\textwidth}
      \includegraphics[width=\textwidth]{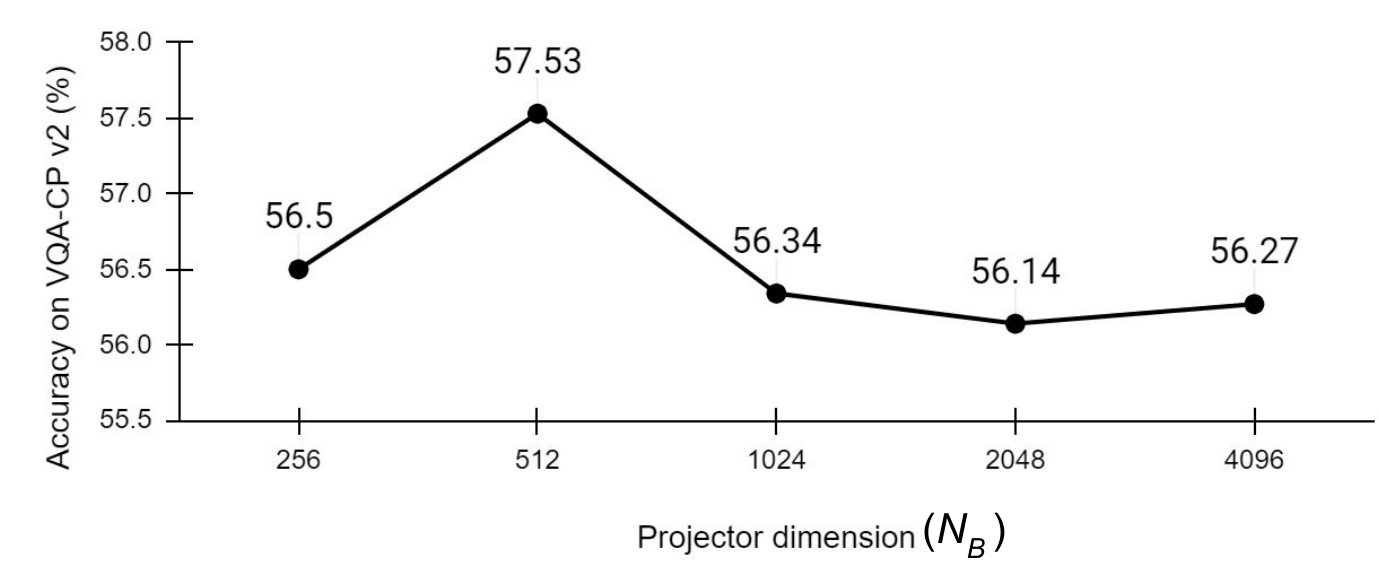}
      \captionof{figure}{\textbf{Effect of the dimensionality of the projector network on the answering performance:} Answering performance on VQA-CP v2 \cite{agrawal2018don} dataset using COB model with different projector dimensions. Best performing model has a projector with output dimensionality $N_B=512$.}
      \label{fig:nb_vs_acc}
  \end{minipage}
    \vspace{-2em}
  \end{minipage}

\subsection{Projector dimensionality selection}
\vspace{-0.3em}
Zbontar and Jing \etal \cite{zbontar2021barlow} show that increase in the projector's ($b_{\theta_B}$) output dimensions ($N_B$) 
improves the input self-supervised feature space.
However, for our Barlow decorrelation constraint we found that larger projection spaces, $\mathcal{D}^{B} \rightarrow \mathbb{R^{N_B}} \text{ for } N_B \in \{1024,2048, 4096\}$, have more redundant components. To analyse this, we compute the PCA eigen values for the representations for higher projection spaces, as shown in Table \ref{tab:PCA_proj}. We observe that top-512 components can preserve $\sim 99\%$ of total energy of the embedding space
and hence we choose $N_B = 512$ as the projection dimension for all three Barlow projectors (i.e. $b_{\theta_{B_M}}, b_{\theta_{B_A}},b_{\theta_{B_{MA}}}$). As shown in Figure \ref{fig:nb_vs_acc}, $N_B = 512$ also yields the maximal answering accuracy compared to other values of $N_B$, which reaffirms our choice. In supplementary, we provide more ablation analysis and pseudo-code for our methods. 
\vspace{-0.7em}
\subsection{Generalizability} \vspace{-0.3em}
Both the ATB and COB models improve the latent representation of a typical classification-based VQA model by imposing a multimodal redundancy minimization constraint on the categorical loss. In sections \ref{sec:atb_ablation} and \ref{sec:qual_results}, we show that ATB and COB models built upon the base GGE outperforms it by learning a more informative latent space, Figure \ref{fig:result_cam}. GGE \cite{han2021greedy} model is the SOTA for VQA, and hence while improving it validates our proposed models, it also raises the question if the improvement in the results only comes due to the better latent features of the base GGE model. In other words, does the improvement in results is dependent on the better quality of the base model. To study this we apply ATB and COB constraints on the UpDn \cite{anderson2018bottom} model, which itself is the base of GGE model. The resulting ATB-UpDn and COB-UpDn models outperform (answering accuracy: $47.36\%$ and $48.24\%$ respectively) the base UpDn model ($39.38\%$) by a significant margin on VQA-CP v2 dataset. This shows that our constraint formulation, despite being limited by the quality of the base model, imposes a regularization on the latent features to be more informative, resulting in an improved performance over the corresponding baseline.

\vspace{-0.7em}
\section{Limitations}\vspace{-0.5em}
\textbf{Lagrange multiplier's initialization:} We initialize $\lambda$ such that it brings both objective loss and constraint loss to the same range. Based on our empirical evidence this scheme works well for different VQA models, however generalization of this scheme for other tasks has not been explored

\textbf{Projector dimensionality selection:} As shown in Figure \ref{fig:nb_vs_acc}, $N_B$ is an important hyperparameter, and is dependent on the input data distribution. We use PCA energy computation to decide this parameter value. However, it is expensive to compute PCA for large datasets and for high dimensional features.

\vspace{-0.5em}
\section{Conclusion}\vspace{-0.5em}
We propose a new VQA regularization scheme called COB that optimizes cross-entropy loss while subjected to a redundancy minimization constraint. Cross-modal Barlow decorrelation loss as the constraint formulation promotes the alignment of the answer with image-and-question modalities while improving the information content of the underlying feature space. We propose two training policies, ATB and COB, to balance these two losses.
We show that both ATB and COB outperform the most recent SOTA GGE baseline, Table \ref{tab:abl}, on VQA-CP v2 and VQA v2 datasets for the answer prediction task. COB model also either outperforms or provides comparative results against other SOTA baselines, Table \ref{sota}
without using additional annotations. Finally, Figure \ref{fig:result_cam} shows that our model focuses more on the salient regions while answering the questions, hence being more interpretable. 

\clearpage
%
%
\bibliographystyle{splncs04}
\bibliography{main}

\appendix
\appendix

\section{Supplementary: Barlow constrained optimization for Visual Question Answering}
In this supplementary, we provide additional details on our proposed models and experiments. In Section \ref{sec:GGE_details}, we discuss the baseline GGE model \cite{han2021greedy} and compare its architecte with our proposed Constrained Optimization with Barlow (COB) model. In Section \ref{sec:implementation_details} we provide the details on the implementation, datasets used for training and evaluation, architectural details, and the hyperparameters. In Section \ref{sec:algo}, we discuss the algorithms for our models. Section \ref{sec:hyper} has a detailed analysis of hyperparameters ($\lambda, \kappa, step$ size). We also extended Section 5.2 from the main paper by providing more details on the selection of Barlow projector's output dimensionality, $N_B$. Additional qualitative and explainability results are provided in Section \ref{sec:add_qual_exp}. Finally, we provide a list of all the mathematical notations used in the main paper and supplementary in the glossary, Section \ref{sec:glossary}.

\vspace{-.7em}

\section{Brief discussion on GGE-DQ-iter Model.}
\label{sec:GGE_details}
\begin{figure}
    \centering
    \includegraphics[width=\linewidth]{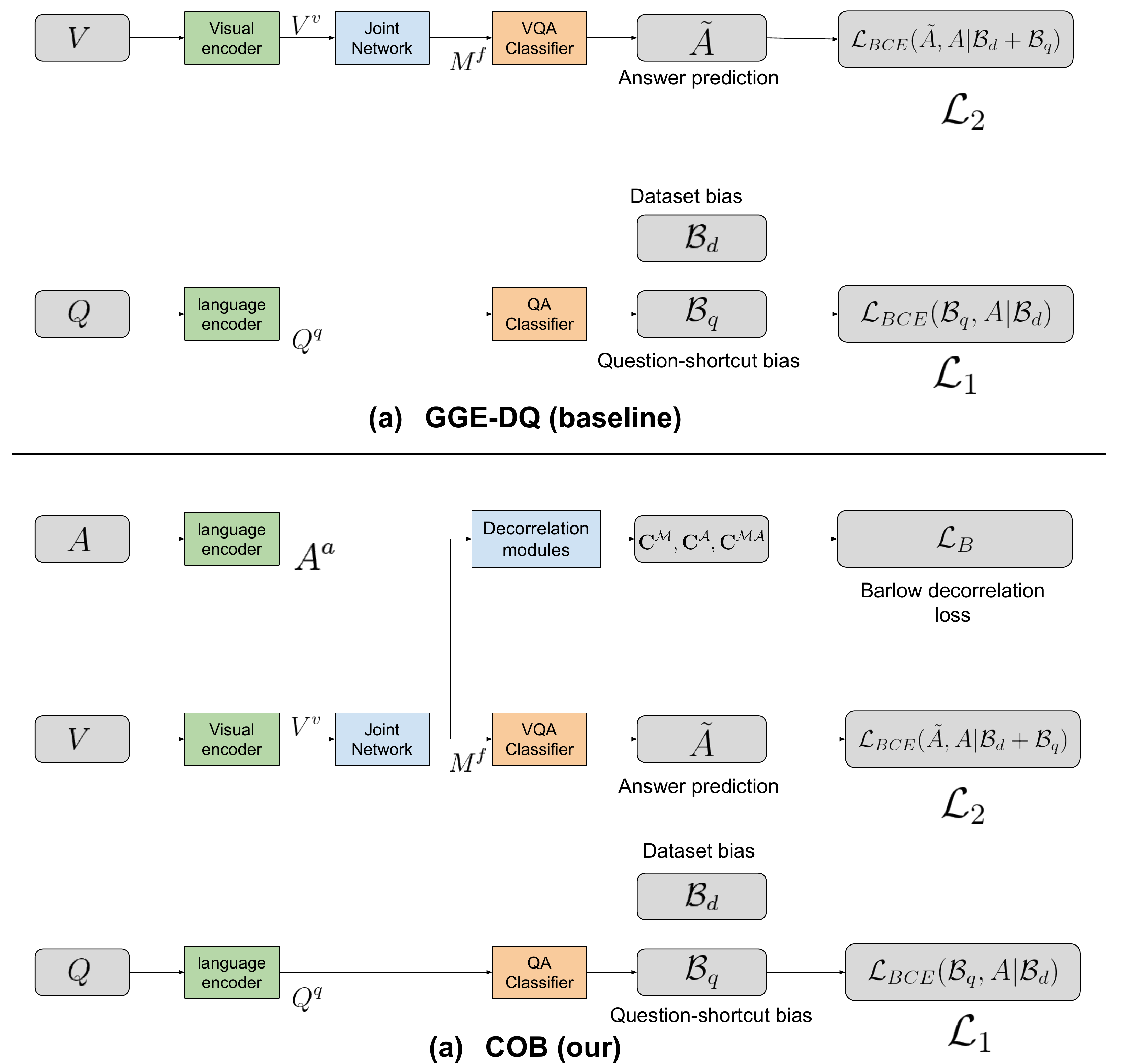} 
      \caption{\textbf{GGE-DQ v/s COB:} (a) shows the baseline GGE-DQ model \cite{han2021greedy}, (b) shows our proposed COB model built upon the base GGE-DQ model. In the main paper, we formulate COB and ATB over the cross-entropy loss $\mathcal{L}_{CE}$ for a generic classification-based VQA model. However, GGE-DQ uses binary cross-entropy(BCE) as the categorical loss. It models two biases in its loss: a distribution bias $\mathcal{B}_d$ and a question-shortcut bias $\mathcal{B}_d$. Conditioned upon these biases, two BCE losses are computed $\mathcal{L}_{1}, \mathcal{L}_{2}$ for the question-only stream and the vision+question stream, respectively. Hence, to build our COB with GGE-DQ as the base architecture, we also use BCE loss, as shown in (b). The constraint formulation and balancing of losses remain the same, as proposed in the main paper, for the generic VQA model. A detailed discussion on the dataset bias and question-shortcut bias can be found in Han \etal \cite{han2021greedy}.}
      \label{fig:COB_GGE_block}
      \vspace{-1.5em}
  \end{figure}

We use GGE-DQ-iter\cite{han2021greedy} as our baseline model.  This model consists of an image encoder and text encoder for image and question input, respectively, similar to the UpDn\cite{anderson2018bottom} architecture. The GGE model uses a self-attention network to get the joint feature representation by combining image encoded and question encoded features. Finally, a classifier network to predict the answer for the given image and question input. The GGE-DQ-iter method uses a two-stage training mechanism to train the model. In the first stage, the model tries to overcome the question bias, and the second stage it tries to overcome the distributional bias in an iterative fashion. Figure \ref{fig:COB_GGE_block} shows the block diagrams for the baseline GGE-DQ model and our COB model built upon this base GGE-DQ architecture.
A detailed analysis of the loss function and exact model details are available in the Han \etal \cite{han2021greedy}.

  \section{Implementation details}
  \label{sec:implementation_details}
   \textbf{Dataset} To evaluate our proposed model, we conduct experiments on the standard VQA v2 \cite{goyal2017making} and language-bias sensitive VQA-CP v2 \cite{agrawal2018don} datasets. VQA v2 dataset contains 443K train, 214K val, and 453K test question-answer pairs corresponding to 83K train, 40K val, and 81K test images sampled from MS COCO datasets. VQA-CP v2 contains the same data as VQA v2 while overcoming its language bias by restructuring the answers and questions in the training and the validation sets, such that prior distribution of answers for every question type in the train and validation set differ from each other. The redistribution of data makes VQA-CP v2 more balanced and robust to language bias.

\textbf{Architecture:} In our model, we use Bottom-Up and Top-Down (UpDn)\cite{anderson2018bottom} features as input for image representation, and GloVe~\cite{pennington2014glove} based word embedding for question tokens input followed by an LSTM~\cite{hochreiter1997long} to obtain Question representation. We use an attention mechanism to combine visual feature and question representation to obtain joint representation, followed by a classifier to obtain answer logits. For each example (consisting of image, question, and answer) in the VQA dataset, we obtain a joint embedding of image-and-question and an answer token embedding based on the GloVe word embedding model. We use a two-stream model between the joint representation and the answer representation. We project the joint representation and the answer representation to a latent embedding space using a projector network, as shown in Figure 2 of the main paper. The projector network has two linear layers, each of dimension 512 output units. The first layer of the network consists of a linear layer followed by a rectified linear unit followed by a second linear layer.  We add the output of the first linear and second layers, followed by a normalization layer to get the final projection embedding. The joint representation is used for the answer prediction task, and the projected embeddings are fed to the Barlow decorrelation loss function.


\begin{figure}
    \centering
    \includegraphics[width=\linewidth]{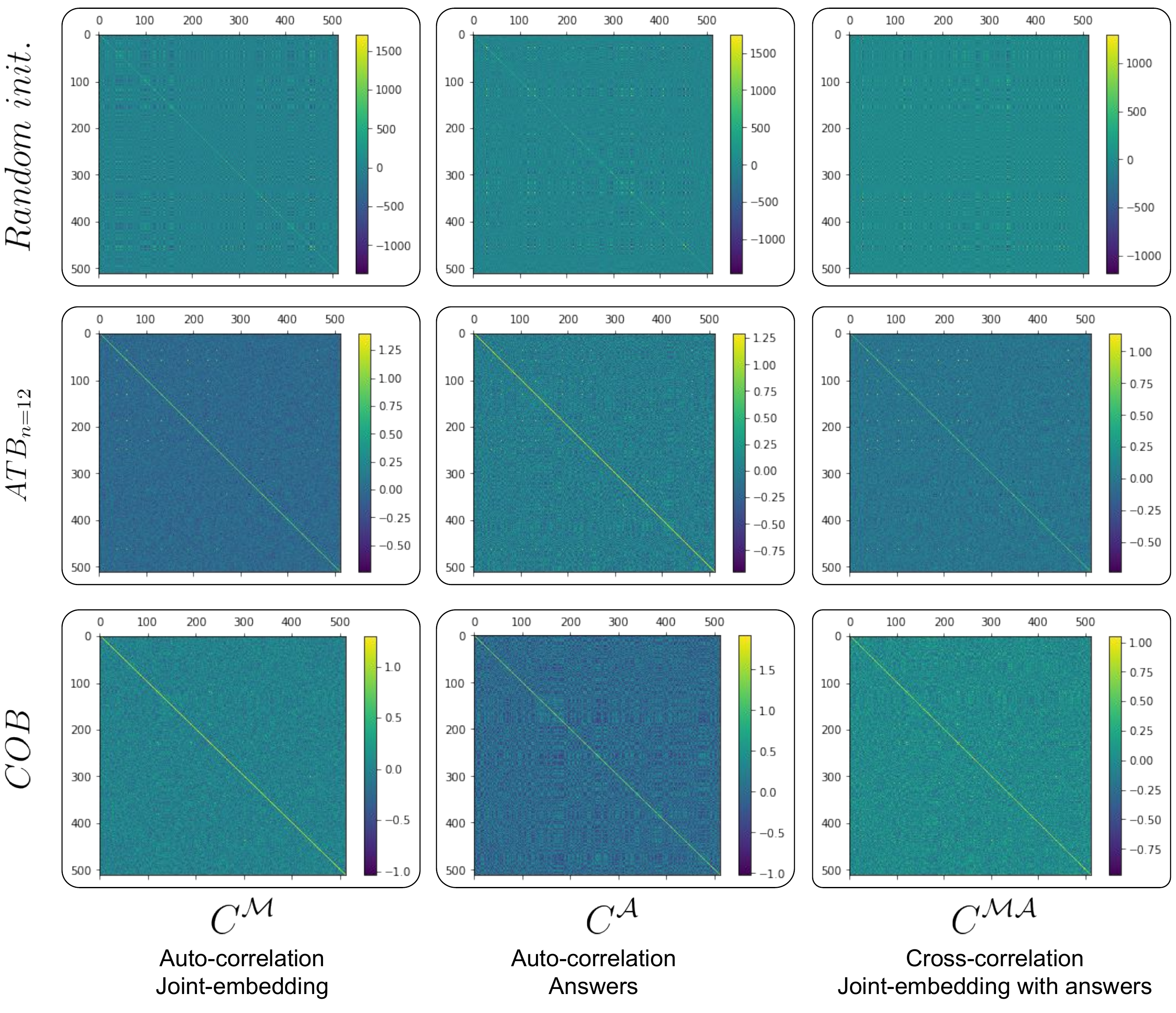} 
      \caption{\textbf{Decorrelation in Barlow space:} Figure shows the auto-correlation and cross-correlation matrices for a randomly initialized network (top-row), $ATB_{n=12}$ model (middle-row) at convergence and COB model (bottom-row) at convergence. Barlow decorrelation loss forces the feature components to share less information with other feature components by decorrelating them, as can be seen by higher value diagonal elements in the auto-correlation matrices ($C_\mathcal{M}$ and $C_\mathcal{A}$) at convergence for both ATB and COB models. Our proposed multimodal Barlow decorrelation loss ($\mathcal{L}^\mathcal{MA}_B$) also forces the two modalities to be aligned along their major component axes while being decorrelated along the feature dimension, as can be seen by a prominent diagonal in the cross-correlation matrix ($C_\mathcal{MA}$).}
      \label{fig:decorr_barlow_space}
      \vspace{-0.5em}
  \end{figure}

\textbf{Decorrelation in Barlow space:}
In Figure \ref{fig:decorr_barlow_space}, we visualize the correlation matrices in the $N_B$ dimensional Barlow space where decorrelation loss ($\mathcal{L}_B$) is computed. We observe that, for a randomly initialized network, the correlation matrix shows a higher redundancy, as shown by the similar values of the diagonal elements as that of the non-diagonal elements, i.e., a non-prominent diagonal for ($C^\mathcal{M},C^\mathcal{A} and C^\mathcal{MA}$). At convergence, both ATB and COB models (middle and bottom rows of Figure \ref{fig:decorr_barlow_space}) show (i) a prominent diagonal in auto-correlation matrices ($C^\mathcal{M},C^\mathcal{A}$), which means the feature components share less information with other feature components and thus being more informative. (ii) A prominent diagonal in cross-correlation matrix ($C^\mathcal{MA}$), implies that our multimodal Barlow decorrelation loss aligns the two modalities (join-embedding and answers) along the major components while keeping the individual feature components decorrelated with each other. This alignment of features between the two modalities helps the underlying joint-embedding to learn the semantics of the answer space (embedded in the GloVe word embedding space), which is otherwise not possible by using only the categorical loss.


\begin{table}
  \centering
  \caption{Ablation analysis: Applying Barlow loss after certain epoch. All the results are \% answering accuracy on VQA-CP v2 test set.}\label{tab:epoch}
  \begin{tabular}{@{}lcccc@{}}
    \toprule
    Method & All & Y/N & Number & Other \\
    \hline 
    baseline (GGE) &\textbf{56.08} & 86.64& 22.15&49.38\\
    $ATB_{n=0}$ & 53.64&87.58 &14.94 &46.47\\
    $ATB_{n=2}$ & 55.19&85.51 & 20.22&48.89\\
    $ATB_{n=4}$ & 55.60 &86.29 & 23.94  &48.21\\
    $ATB_{n=6}$ & 56.75& 87.57&22.82 &49.90\\
    $ATB_{n=8}$ &56.76 &87.81 &23.08 &49.73\\
    $ATB_{n=10}$ & 56.74& 87.70&23.14 &49.73\\
    $ATB_{n=11}$ & 57.16 & 87.34&\textbf{27.45} &\textbf{49.53}\\
    $ATB_{n=12}$ & \textbf{57.18} & \textbf{87.53} & 27.19 & 49.51\\
    $ATB_{n=13}$ & 56.80&87.71 &24.34 &49.50\\
    $ATB_{n=14}$ & 56.77&87.62 &23.85 &49.53\\
    $ATB_{n=16}$ &56.59 &87.10 &23.90 &49.58\\
    $ATB_{n=18}$ &56.38 &87.94 &20.98 &49.57\\ 
    \bottomrule
  \end{tabular}
 \end{table}

\textbf{Ananlysis on the amount of pretraining for ATB:}
Here, we extend Section 4.2 of the main paper. In Table \ref{tab:epoch}, we present additional VQA results using our ATB model for different pretraining epochs, \(n\). We evaluate on different types of questions sets from the VQA-CP v2 \cite{agrawal2018don} dataset, namely: Y/N, Number, and Other, along with the overall results on all of these sets.

\subsection{Algorithms}
\label{sec:algo}
To elaborate the formulation and training policies for the proposed ATB and COB models, we provide the respective algorithms in \textit{Alorithm} \ref{alg:atb} and \ref{alg:cob}. All the mathematical notations are defined in Section 3 and visually placed in Figure 2. We also provide a glossary of all the notations in Section \ref{sec:glossary}.

\SetKwComment{Comment}{/* }{ */}

\SetKwInOut{KwInput}{Input}
\SetKwInOut{KwResult}{Result}
\SetKwInOut{KwParameters}{Parameters}

\begin{algorithm}
\SetAlgoLined
\caption{Align then Barlow (ATB)}\label{alg:atb}
\KwInput{Batches (V,Q,A),\(n\)}
\KwParameters{\(\theta_J, \theta_L \text{ and }  \theta_B = \{\theta_{B_M}, \theta_{B_A}, \theta_{B_{MA}}\}\)}
\KwResult{Learned parameters \(\theta_J, \theta_L, \theta_B\).}
Initialize \(epoch = 0\);\\

\While{is training}{
Compute categorical loss for the current batch, \(\mathcal{L}_{CE}\);\\
\eIf{\(epoch\leq n\)}{
Compute gradients \(G_{\theta_{J}} \leftarrow \frac{\partial\mathcal{L}_{CE}}{\partial\theta_{J}}\) and \(G_{\theta_{L}} \leftarrow \frac{\partial\mathcal{L}_{CE}}{\partial\theta_{L}}\);\\
Update parameters as \(\triangle_{\theta_{J},\theta_{L}} \propto -G_{\theta_{J},\theta_{L}};\)\\
}{
Compute Barlow decorrelation loss for the current batch, \(\mathcal{L}_{B}\);\\
\(\mathcal{L}_{ATB} \leftarrow \frac{1}{2}(\mathcal{L}_{CE}+\mathcal{L}_{B})\);\\
Compute gradients 
\(G_{\theta_{J}} \leftarrow \frac{\partial\mathcal{L}_{ATB}}{\partial\theta_{J}}\),
\(G_{\theta_{L}} \leftarrow \frac{\partial\mathcal{L}_{ATB}}{\partial\theta_{L}}\)
and
\(G_{\theta_{B}} \leftarrow \frac{\partial\mathcal{L}_{ATB}}{\partial\theta_{B}}\)
;\\
Update parameters as:\\ \(\triangle_{\theta_{J}, \theta_{L}, \theta_{B}} \propto -G_{\theta_{J}, \theta_{L}, \theta_{L}};\)\\
}
\(epoch \leftarrow epoch+1;\)\\
}
\end{algorithm}
\vspace{-1.3em}

\SetKwComment{Comment}{/* }{ */}

\SetKwInOut{KwInput}{Input}
\SetKwInOut{KwResult}{Result}
\SetKwInOut{KwParameters}{Parameters}

\begin{algorithm}
\SetAlgoLined
\DontPrintSemicolon
\caption{Constrained optimization with Barlow (COB)}\label{alg:cob}
\KwInput{Batches (V,Q,A), \(\kappa, step\)}
\KwParameters{\(\theta = \{\theta_J, \theta_L, \theta_{B_M}, \theta_{B_A}, \theta_{B_{MA}}\}\), Lagrange multiplier \(\lambda\)}
\KwResult{Learned parameters \(\theta\), and Lagrange multipliers \(\lambda\).}
Initialize \(t = 0, \lambda = 1e-5\);

\While{is training}{
Compute categorical loss for the current batch, \(\mathcal{L}_{CE}\);

Compute the constraint loss over the batch, \(\mathbb{C}^t \leftarrow(\mathcal{L}_{B}-\kappa\));

Compute gradient \(G_\theta \leftarrow \frac{\partial\mathcal{L}_{all_{COB^\lambda}}}{\partial\theta}\);

Update parameters as \(\triangle_\theta \propto -G_\theta\);

Update \(\mathbb{C}^t\), as:\tcp*[l]{Required to compute \(\lambda_{t+1}\)}
\eIf{t == 0}{
\(\mathbb{C}^t \leftarrow(\mathcal{L}_{B}-\kappa)\);

}{
\(\mathbb{C}^t \leftarrow \alpha\mathbb{C}^{t-1} + (1-\alpha)(\mathcal{L}_{B}-\kappa)\);

}

Update \(\lambda\), as:

\eIf{\(t \% step == 0\)}{
  
  \(\lambda_{t+1} \leftarrow \lambda_t\exp(StopGradient(\mathbb{C}^t))\);\tcp*[r]{Update \(\lambda\)}
  }{
  \(\lambda_{t+1} \leftarrow \lambda_t\);
  
    }
  \(t \leftarrow t+1\);
  
}
\end{algorithm}

\begin{figure}
    \centering
    \includegraphics[width=0.8\linewidth]{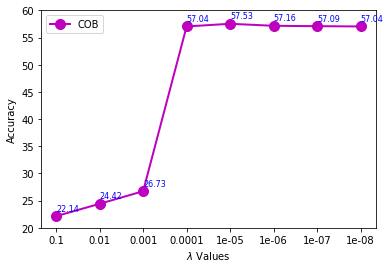} 
    \vspace{-2.5em}
      \caption{Tuning for hyperparameter Lagrange multiplier \(\lambda\).}
      \label{fig:lambda}
      \vspace{-1em}
  \end{figure}
\subsection{ Hyperparameter Selection}
\label{sec:hyper}
\textbf{Selection of \(\lambda\) :}
We perform our experiment for different values of \(\lambda\). We observe that for the large value of \(\lambda\) the loss function does not converge, and for a small value of \(\lambda\), the loss function converges at its optimum performance. We start \(\lambda\) value from 0.1  and increased in its order of magnitude up to \(\lambda = 0.00000001\) value. We observe that for \(\lambda= 0.00001\), the model performs best out of all other values of \(\lambda\), as shown in the Figure \ref{fig:lambda}. 
\begin{figure}
\vspace*{-0.5cm}
    \centering
    \includegraphics[width=0.8\linewidth]{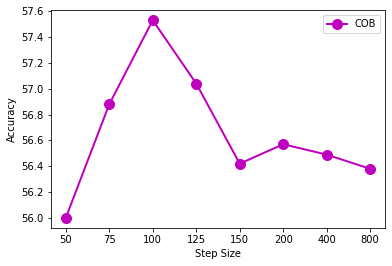} 
    \vspace{-2em}
      \caption{Tuning for hyperparameter Step size.}
      \label{fig:step}
      \vspace{-1em}
  \end{figure}
  
\textbf{Selection of \(step\) size :} 
The tuning of the hyperparameter, step size,  (Number of Iteration)  plays a crucial role in training our COB model. The \(\lambda\) value updates after a specific step size. We perform our experiment with different values of step sizes (Number of Iteration) starting from step size 50 to step size 800 as shown in Figure \ref{fig:step}. We observe that step size 100 performs better than other step sizes. 
  
\begin{figure}
    \centering
    \includegraphics[width=0.8\linewidth]{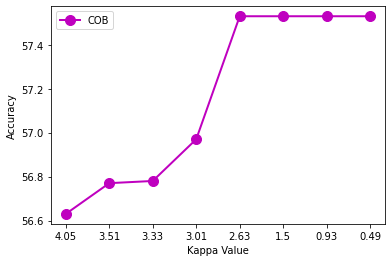} 
    \vspace{-0.7em}
      \caption{Tuning for hyperparameter \(kappa\).}
      \label{fig:kappa}
      \vspace{-1.7em}
  \end{figure}

\textbf{Selection of \(\kappa\) :}
\(\kappa\) is the threshold value which controls when the lambda value starts decreasing. When the \(\kappa\) value reaches the Barlow twin loss value, the constraint value becomes zero, and after that, the constraint becomes negative. The negative constraint value tries to reduce the contribution of Barlow twin loss in the total loss value. The selection of \(\kappa\) value is a complex task. We need to observe the pre-trained model and set the \(\kappa\) value to the saturation value of the Barlow loss( where Barlow loss does not change much). Set the \(\kappa\) value near to that saturation value. We experimented with analyzing the behaviors of \(\kappa\) we set with higher saturation value and lower saturation value as shown in Figure \ref{fig:kappa}.
Based on the empirical observation we select \(\kappa = 2.63\) for training COB model.
We observe, for lower saturation value, the performance does not affect much.

\begin{figure}
    \centering
    \includegraphics[width=0.8\linewidth]{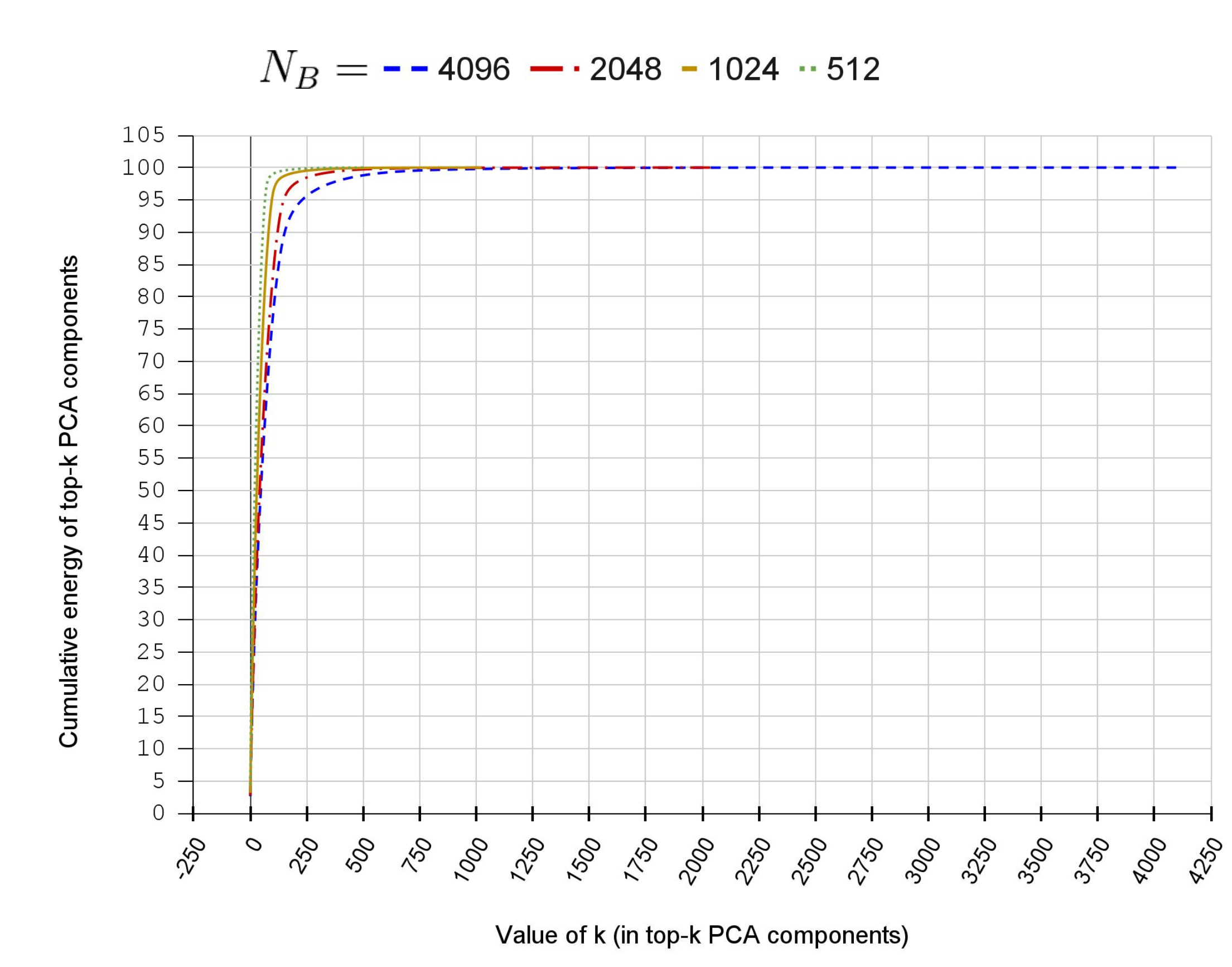} 
      \caption{Cumulative energy of top-k PCA components for different values of Barlow projection layer's (Barlow projector's) output dimensionality.}
      \label{fig:pca_full}
  \end{figure}

\textbf{Selection of \(N_B\) :} \(N_B\) is the output dimensionality of the Barlow projectors (\(b_{\theta_{B_M}},b_{\theta_{B_A}},b_{\theta_{B_{MA}}}\)). It is an important hyperparameter, as a too-small value to \(N_B\) leads to a smaller Barlow space where the multiple semantic concepts would be required to be modelled by the same feature component, and a too-large value would cause multiple feature components to model the same semantic concept. These two cases result in inferior performances, as shown in Figure 7 in the main paper. To select a good value of \(N_B\) we use a PCA analysis. We compute the cumulative energy of the top-k eigenvectors on a subset of VQA-CP v2 test set using our COB model for different values of projector dimension (\(N_B\)). From Figure \ref{fig:pca_full}, we observe that 512 eigenvectors contains at least 98.8\% of total PCA energy for \(N_B \leq 4096\). Hence, we chose \(N_B = 512\) for all our experiments. Figure \ref{fig:pca_full} supplements the Table 3 in the main paper.

\begin{figure}
    \centering
    \includegraphics[width=\linewidth]{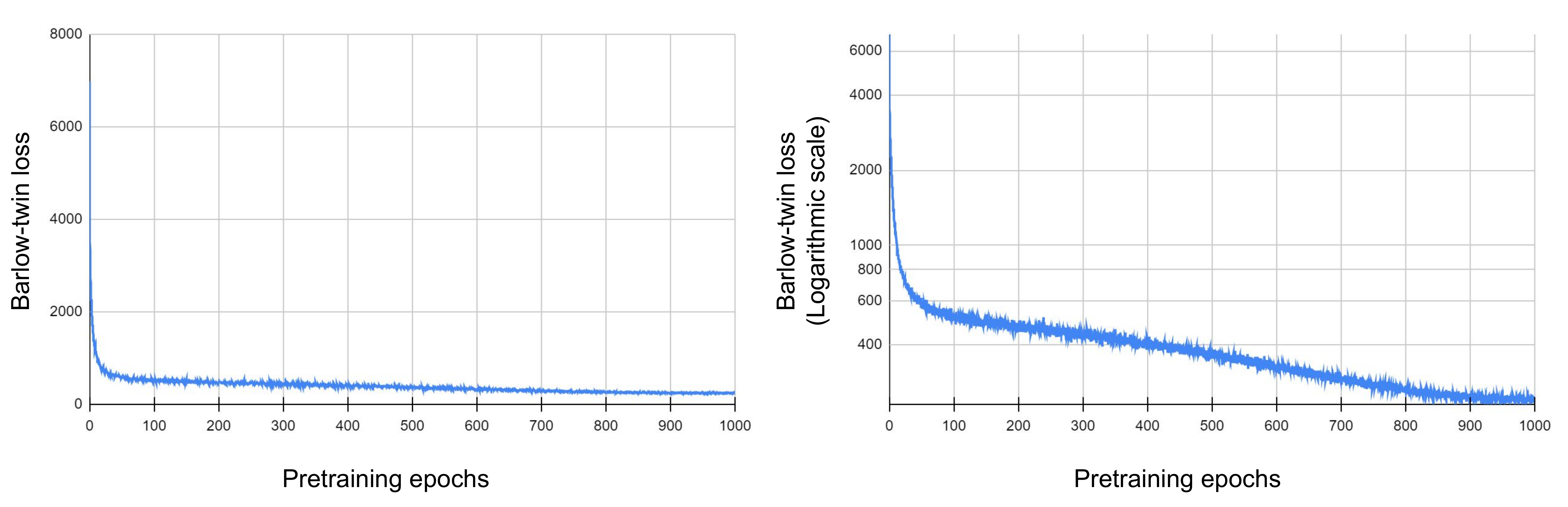} 
      \caption{\textbf{Pretraining loss curve for official implementation of Barlow-twins \cite{zbontar2021barlow}}: (Left) shows the decorrelation loss during pretraining for each epoch, (right) shows the decorrelation loss on a logarithmic scale for each epoch for a better visualization of the flatter region. We observe that Barlow-loss takes a longer gradient cycle to converge. \textbf{Note:} To plot these curves, we use the logs provided by the official implementation of Barlow-twins.}
      \label{fig:orig_barlow_twins}
      \vspace{-1.5em}
  \end{figure}

 \textbf{Analysis of original Barlow-twins loss:}
 In section 3.4.a of the main paper, we discuss that Barlow-twins \cite{zbontar2021barlow} uses 1000 epochs of pretraining, suggesting a flatter loss curve. Here we plot the pre-training loss curve using the logs of the official implementation\footnote{Official implementation of Barlow-Twins \cite{zbontar2021barlow}\\: https://github.com/facebookresearch/barlowtwins} of the Barlow-twins for reference, shown in Figure \ref{fig:orig_barlow_twins}.
  
 \section{Additional Qualitative and Explainability results:}
  \label{sec:add_qual_exp}
  \vspace{-0.3em}
  
\begin{figure*}
    \centering
    \includegraphics[width=\linewidth]{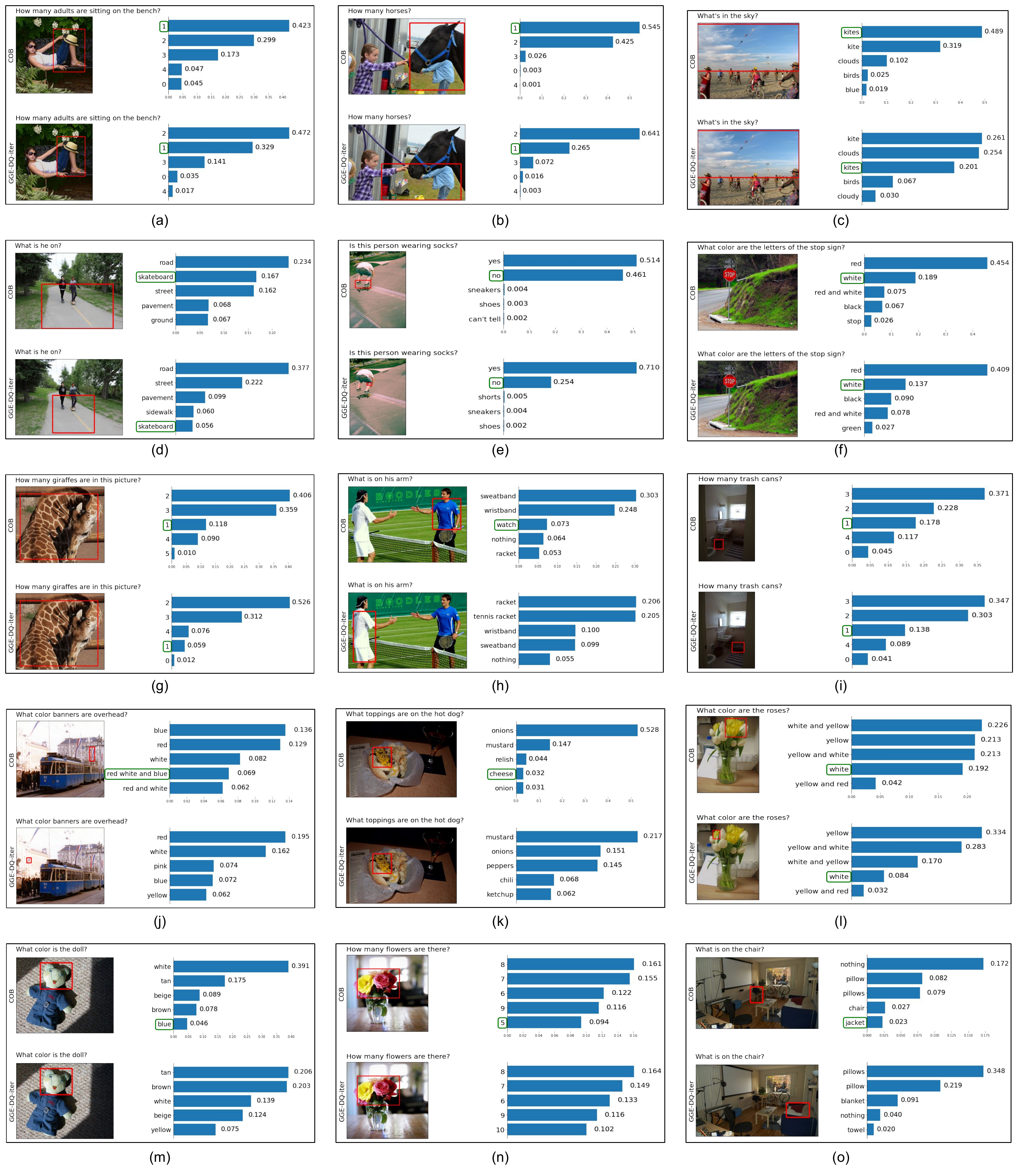} 

      \caption{\textbf{More qualitative results:} Here we extend the qualitative result section of the main paper. Each of the image set/cell shows results for COB model (top-left), with top-5 prediction along with the probability scores corresponding to them, similarly bottom-left shows the GGE-DQ-iter baseline model prediction and bottom-right shows the top-5 baseline predictions with their probability scores. The ground truth answer is denoted by the answer with encapsulating green box. Red bounding box shows the maximal attention region in each image.}
      \label{fig:qualitative_result}
\end{figure*}

\begin{figure*}
    \centering
    \includegraphics[width=\linewidth]{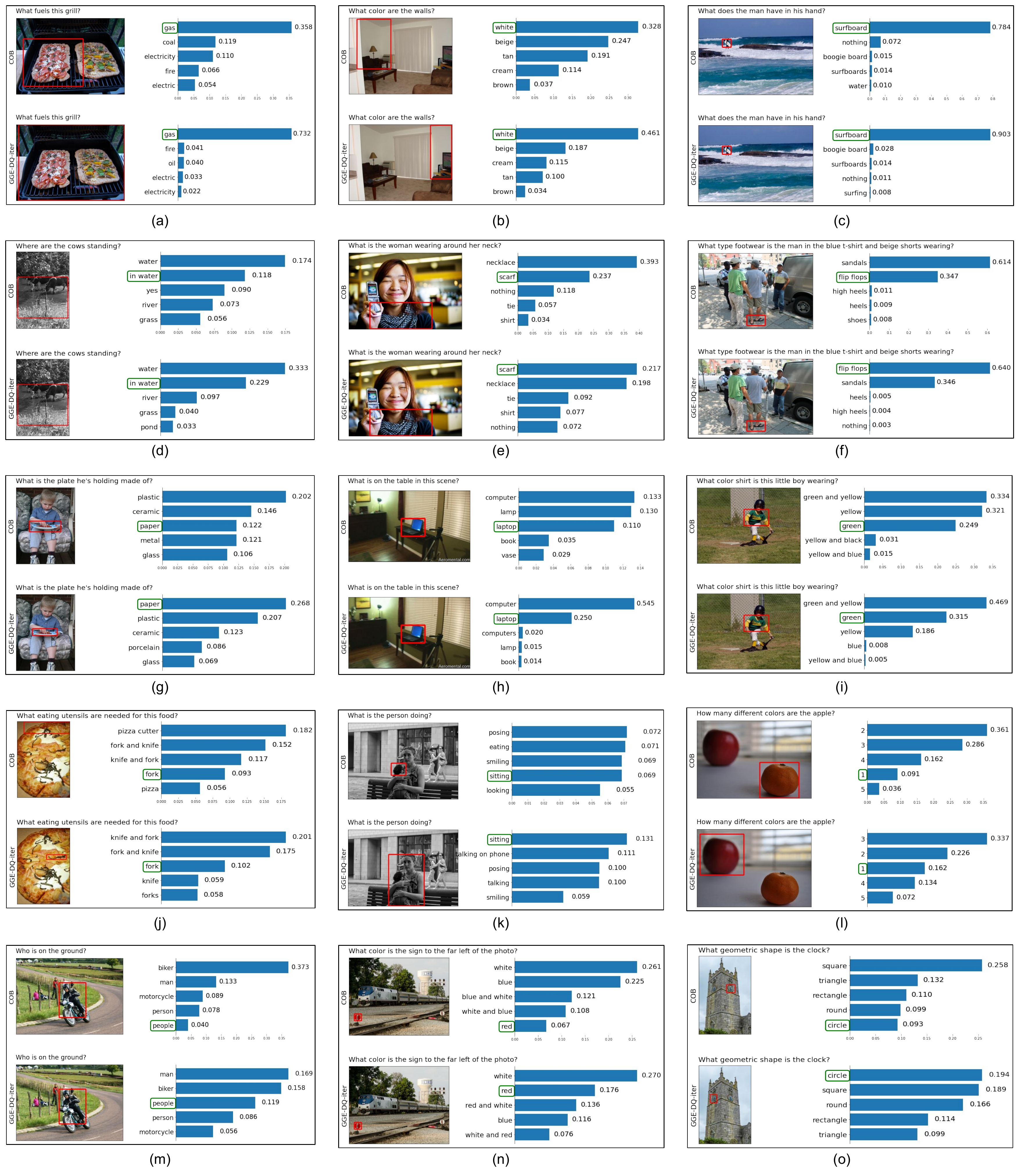} 

      \caption{\textbf{Similar or Negative results w.r.t. baseline model:} Here we explicitly show the results where COB performs either equal or inferior to the baseline model. Each of the image set/cell shows results for COB model (top-left), with top-5 prediction along with the probability scores corresponding to them, similarly bottom-left shows the GGE-DQ-iter baseline model prediction and bottom-right shows the top-5 baseline predictions with their probability scores. The ground truth answer is denoted by the answer with encapsulating green box. Red bounding box shows the maximal attention region in each image.}
      \label{fig:qualitative_result_neg}
\end{figure*}

\begin{figure*}
    \centering
    \includegraphics[width=\linewidth]{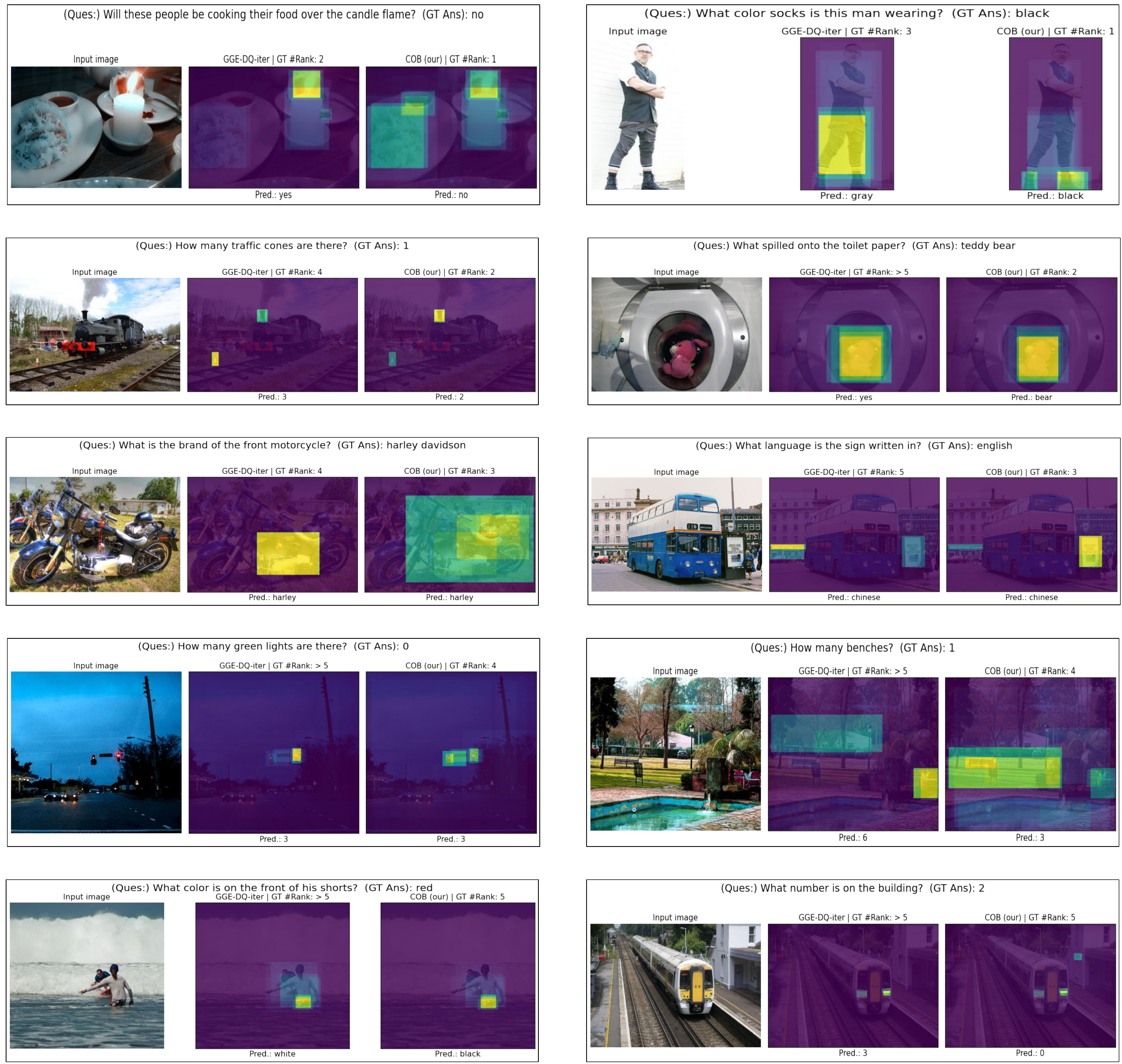} 

      \caption{\textbf{More explainability results:} Here we extend the explainability results of the main paper. For each image set/cell: (top-text) is the input question along with the ground truth (GT) answer; left-image is the input image middle-image is the Grad-CAM \cite{Selvaraju_2017_ICCV} heatmaps computed by the baseline GGE-DQ-iter model overlaid on the original image; right-image is the overlaid Grad-CAM heatmap computed by COB;  GT \#Rank denotes the rank of the ground truth answer in the top-5 prediction by the respective models. `Pred.' at the bottom of the middle and right images denotes the predicted answer with the highest probability score by the respective models.}
      \label{fig:grad_cam_result}
\end{figure*}

\begin{figure*}
    \centering
    \includegraphics[width=\linewidth]{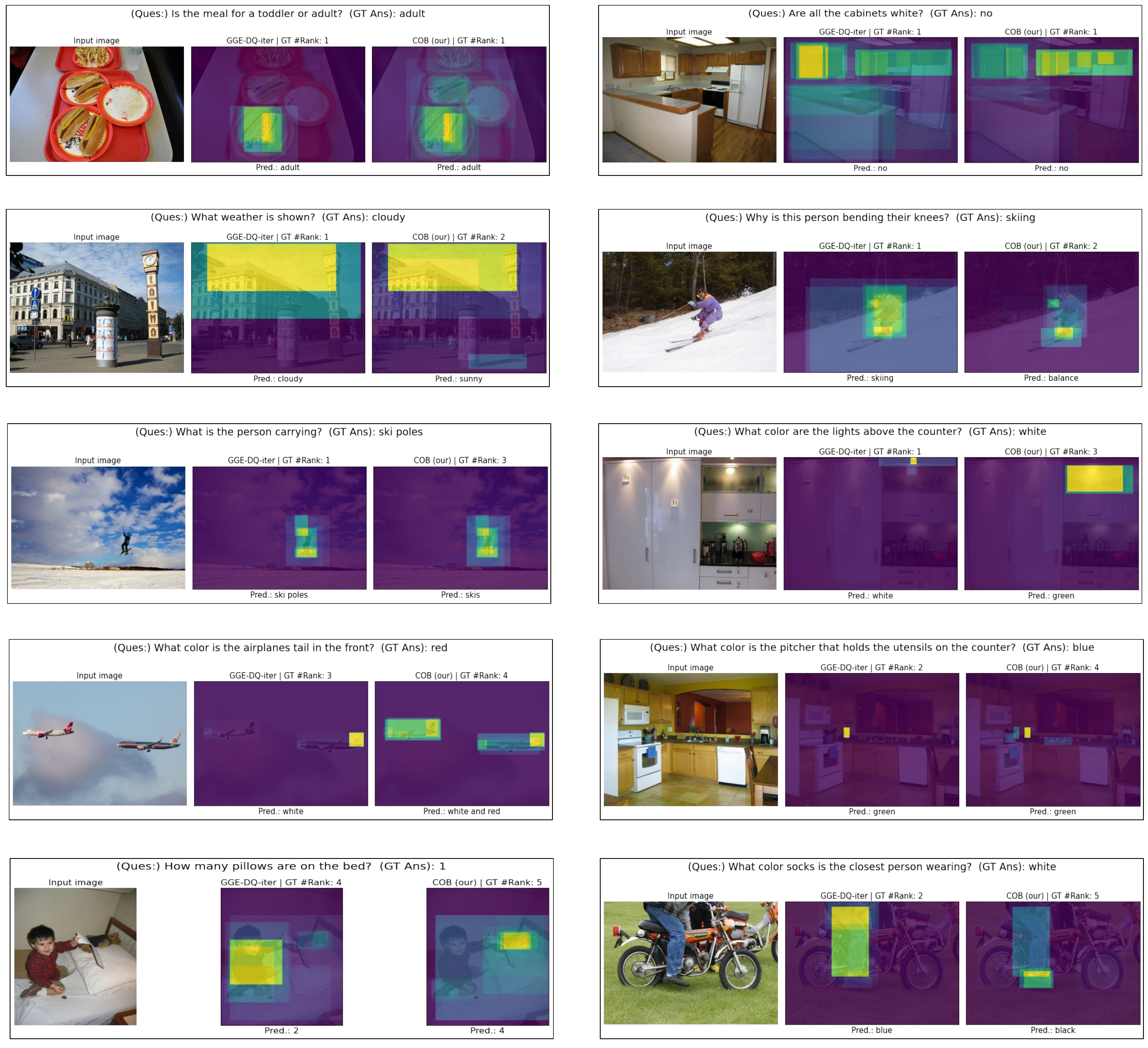} 

      \caption{\textbf{Explainability results when COB performs similar or inferior to baseline (GGE-DQ-iter) model:} We observe that for the cases where the COB performs inferior to the baseline, the COB model still localizes either the same salient regions or better. This property of better salient localization also results in an improved CGD scores obtained by COB in comparison to all other state-of-the-art baselines, as discussed in the main paper.
      For each image set/cell: (top-text) is the input question along with the ground truth (GT) answer; left-image is the input image middle-image is the Grad-CAM \cite{Selvaraju_2017_ICCV} heatmaps computed by the baseline GGE-DQ-iter model overlaid on the original image; right-image is the overlaid Grad-CAM heatmap computed by COB;  GT \#Rank denotes the rank of the ground truth answer in the top-5 prediction by the respective models. `Pred.' at the bottom of the middle and right images denotes the predicted answer with the highest probability score by the respective models.}
      \label{fig:grad_cam_result_neg}
\end{figure*}

  We provide more qualitative results in Figure \ref{fig:qualitative_result}, along with an additional set of inferior performing results in Figure \ref{fig:qualitative_result_neg}. Similarly, more results on explainability using Grad-CAM \cite{Selvaraju_2017_ICCV} have been provided in Figure \ref{fig:grad_cam_result}, along with additional set of results for the cases where COB performs similar or inferior to baseline GGE model in Figure \ref{fig:grad_cam_result_neg}.
  
 \subsection{Glossary}
\label{sec:glossary}
A glossary of all the mathematical notation used in the main paper and supplementary can be found in Table \ref{tab:gloss_1}, \ref{tab:gloss_2}.
  
\clearpage
\vfill
\newpage

\clearpage
\vfill
\newpage
  
\clearpage
\vfill
\newpage

\begin{table*}[hb]

   \caption{\textbf{Glossary of notations:} Definition of the notation use in the main manuscript and supplementary. }
  \label{tab:gloss_1}
  \centering
\begin{tabular}[t]{@{}l p{9.5cm}@{}}
    \toprule
    Notation & Meaning \\
    \midrule
    \(\mathcal{D}^{VQA}\)  & Distribution of input image question and answers.\\
    \(\mathcal{D}^{V}\) & Distribution of input images. \\
    \(\mathcal{D}^{Q}\) & Distribution of input question. \\
    \(\mathcal{D}^{A}\) & Distribution of input answers. \\
    \(d_k\) & An instance sampled from \(\mathcal{D}^{VQA}\), indexed by \(k\). \\
    \(v_k\) & An instance sampled from \(\mathcal{D}^{V}\). \\
    \(q_k\) & An instance sampled from \(\mathcal{D}^{Q}\). \\
    \(a_k\) & An instance sampled from \(\mathcal{D}^{A}\). \\
    \(n_b\) & Number of samples in a mini-batch.\\
    \(V\) & A mini-batch of \(n_b\) different instances (\(v_k\)) sampled from \(\mathcal{D}^{V}\).\\
    \(Q\) & A mini-batch of \(n_b\) different instances (\(q_k\)) sampled from \(\mathcal{D}^{Q}\).\\
    \(A\) & A mini-batch of \(n_b\) different instances (\(a_k\)) sampled from \(\mathcal{D}^{A}\).\\
    \(e_v\) & Pretrained image encoder, parameters not updated during training.\\
    \(e_q\) & Pretrained language encoder, parameters not updated during training.\\
    \({f}_{\theta_J}\) & Joint network with learnable parameters \(\theta_J\)\\
    \(m^f_k\) & A sample in the joint image+question embedding space.\\
    \(M^f\) & A mini-batch of \(n_b\) different instances (\(m^f_k\)) sampled from \(\mathcal{D}^{M}\).\\
    \(\mathcal{D}_M\) & Distribution of samples (\(m^f_k\)) in the joint embedding space.\\
    \(l_{\theta_L}\) & A non-linear projection layer from joint embedding space to answer logit space.\\
    \(m^l_k\) & Predicted answer logits.\\
    \(M^l\) & A mini-batch consisting of \(n_b\) different instances of (\(m^l_k\)).\\
    \(\mathcal{L}_{CE}\) & Cross-entropy loss, in general Categorical loss. For GGE \cite{han2021greedy}, it is binary-cross entropy loss.\\
    \(N_B\) & Dimensonality of space in which Barlow decorrelation loss is computed.\\
    \(I\) & Identity matrix in real-value space (\(\mathbb{R}\)), of size (\(N_B \times N_B\)).\\
    \(\mathcal{D}^{B}\times \mathcal{D}^{B}\) & A distribution space of matrices \(C\) computed between samples in \(\mathcal{D}^{B}\).\\
    \(\mathcal{D}^S\) & a modality specific distribution: For answers and joint representations, it is \(\mathcal{D}^{A}\) and \(\mathcal{D}^{M}\) respectively.\\
    \(s_k\) & An instance sampled from \(\mathcal{D}^S\).\\
    \(S\) &  A mini-batch of \(n_b\) different instances (\(s_k\)) sampled from \(\mathcal{D}^{S}\).\\
    \(e_{s}\) & Modality specific encoder. For questions, answers and images it is \(e_q, e_a\) and \(e_v\) respectively.\\
    
    \bottomrule

\end{tabular}
\end{table*}
\begin{table*}[hb]

   \caption{\textbf{Glossary of notations:} Definition of the notation use in the main manuscript and supplementary. (Continuation of table \ref{tab:gloss_1}.)}
  \label{tab:gloss_1_1}
  \centering
\begin{tabular}[t]{@{}l p{8.5cm}@{}}

\toprule
    Notation & Meaning \\
    \midrule
    \(s_{k_1} \text{and} s_{k_2}\) & Two complementary samples \(k={k_1,k_2}\) sampled from \(\mathcal{D}^S\), that makes a positive pair. In Barlow twin \cite{zbontar2021barlow}, these are two different augmentations of the same image.\\
    \(s^s_k\) & Encoded representation of \(s_k\) using the encoder \(e_{s}(.)\)\\
    \(S^s\) &  A mini-batch consisting of \(n_b\) different instances of (\(s^s_k\)).\\
    \(b_{\theta_B}\) & A non-linear projector from encoded representation space \(e_{s}(s_k)\) to Barlow space, parameterized by learnable parameters \(\theta_B\).\\
    \(s^b_k\) & A non-linear projection of encoded representation \(e_{s}(s_k)\) in the Barlow space.\\
    \(S^b\) & A mini-batch consisting of \(n_b\) different instances of (\(s^b_k\)). \\
    \(S^{b}_{1} and S^{b}_{2}\) & Two complementary batches consisting of positive pairs.\\
    \(Norm_b(.)\) & Batch normalization function \cite{ioffe2015batch}.\\
    \(C(.)\) & Correlation between two batches.\\
    \(C^S\) & Correlation matrix between two complementary batches \(S^{b}_{1} and S^{b}_{2}\).\\
    \(i and j\) & \(i\) and \(j\) indexes the different feature components of the 
 projected feature vector \(s^b_k\).\\
    \(C^S_{ij}\) & A single element of the correlation matrix \(C^S\) indexed by (\(i,j\)).\\
    \(\mathcal{L}^\mathcal{S}_{B}\) & Barlow decorrelation loss  for unimodal \(\mathcal{D}^S\) input space.\\
    \(e_a\) & Pretrained language encoder, parameters not updated during training.\\
    \(A^a\) & Encoded answer representation for the mini-batch \(A\), using answer encoder \(e_a(.)\).\\
    \(V^v\) & Encoded image representation for the mini-batch \(A\), using answer encoder \(e_v(.)\).\\
    \(Q^q\) & Encoded question representation for the mini-batch \(A\), using answer encoder \(e_q(.)\).\\
    \(b_{\theta_{B_M}}\) & A non-linear projector from the joint representation space \(M^l \in \mathcal{D}^{M}\) to Barlow space, parameterized by learnable parameters \(\theta_{B_M}\).\\
    \(C^\mathcal{M}\) & Auto-correlation matrix computed on the barlow projection (\(b_{\theta_{B_M}}(.)\)) of the batch (\(M^l\)).\\
    \(b_{\theta_{B_A}}\) & A non-linear projector from the encoded image representations \(A^a\) to Barlow space, parameterized by learnable parameters \(\theta_{B_A}\).\\
    \(C^\mathcal{A}\) & Auto-correlation matrix computed on the barlow projection (\(b_{\theta_{B_A}}(.)\)) of the batch (\(A^a\)).\\
    \(C^\mathcal{MA}\) & Cross-correlation matrix computed between barlow projected joint-representations and the encoded answer representations.\\
    
    \(\mathcal{L}^\mathcal{O}_{B}\) & A Barlow decorrelation loss, where \(\mathcal{O}\) denotes the input modalities.\\
    \bottomrule

  \end{tabular}

\end{table*}

\clearpage
\vfill
\newpage

\begin{table*}[hbt]
\caption{\textbf{Glossary of notations:} Definition of the notations used in the main manuscript and supplementary. (Continuation of table \ref{tab:gloss_1}.) }
  \label{tab:gloss_2}
\centering
  \begin{tabular}[t]{@{}l p{8.5cm}@{}}
\toprule
    Notation & Meaning \\
    \midrule
    \(\mathcal{L}^\mathcal{M}_{B}\) & A unimodal Barlow decorrelation loss  for joint image-question embedding space (\(\mathcal{D}^M\)).\\
    
    \(\mathcal{L}^\mathcal{A}_{B}\) & A unimodal Barlow decorrelation loss  for answer space (\(\mathcal{D}^A\)).\\
    \(\mathcal{L}^\mathcal{MA}_{B}\) & A multimodal Barlow decorrelation loss between the joint image-question embedding space (\(\mathcal{D}^M\)) and  answer space (\(\mathcal{D}^A\)).\\
    
    \(\mathcal{L}_{B}\) & Overall Barlow decorrelation loss.\\
    \(\mathcal{L}_{all_{base}}\) & Baseline (naive) implementation of overall (categorical \(+\) Barlow decorrelation) losses.\\
    \(n\) & Number of pretraining epochs (with categorical loss \(\mathcal{L}_{CE}\)) before applying barlow decorrelatin loss (\(\mathcal{L}_{B}\)), in Align then Barlow (ATB) formulation.\\
    \(\mathcal{L}_{all_{ATB}}\) & Overall loss formulation for ATB training policy.\\
    \(\mathcal{L}_{all_{COB}}\) & Our overall constrained optimization formulation (i.e. categorical loss constrained with Barlow (COB) decorrelation loss formulation).\\

    
    \(\lambda\) & A learnable Lagrange multiplier to weight categorical loss \(\mathcal{L}_{CE}\)  and Barlow decorrelation loss \(\mathcal{L}_{B}\).\\
    \(\kappa\) & It is a tolerance hyperparameter to control the change in \(\lambda\), give the value of Barlow decorrelation loss \(\mathcal{L}_{B}\) at iteration \(t\).\\
    \(\mathbb{C}\) & Barlow constraint defined as difference between \(\mathcal{L}_{B}\) and \(\kappa\). When it becomes zero the change in \(\lambda\) becomes negative. This subsequently forces the dynamic weigth \(\lambda\) assigned to constraint to decrease.\\
    \(\mathbb{C}_t\) & Constraint \(\mathbb{C}\) at iteration \(t\).\\
    \(\lambda_t\) & value of \(\lambda\) at iteration \(t\).\\
    \(\mathcal{L}_{all_{COB^\lambda}}\) & Lagrangian form of overall constrained optimization \(\mathcal{L}_{all_{COB}}\).\\
    \(\triangle\lambda_t\) & Change in \(\lambda\) in iteration \(t\).\\

    \bottomrule

  \end{tabular}
 \vspace{-1.2em}
\end{table*}

\clearpage
\vfill
\newpage

\end{document}